\documentclass[11pt]{article}

\usepackage[letterpaper, margin=1in]{geometry}
\usepackage{microtype}
\usepackage[skip=6pt plus 1pt, indent=0pt]{parskip}

\usepackage[T1]{fontenc}
\usepackage[utf8]{inputenc}

\usepackage{lmodern}

\usepackage{amsmath, amssymb, amsthm, mathtools}
\usepackage{bm}

\usepackage{xcolor}

\definecolor{titleboxbg}{RGB}{117, 123, 44}     

\definecolor{linkcolor}{RGB}{188, 196, 195}

\definecolor{rulegray}{RGB}{80, 95, 95}

\definecolor{headingcolor}{RGB}{31, 42, 46}

\definecolor{codebg}{RGB}{248, 248, 248}
\definecolor{codecomment}{RGB}{80, 130, 80}
\definecolor{codekeyword}{RGB}{30, 30, 180}
\definecolor{codestring}{RGB}{160, 40, 40}

\newtheoremstyle{sansplain}      
  {\topsep}{\topsep}{\itshape}{}
  {\sffamily\bfseries\color{headingcolor}}{.}{ }{}
\newtheoremstyle{sansdef}        
  {\topsep}{\topsep}{\normalfont}{}
  {\sffamily\bfseries\color{headingcolor}}{.}{ }{}
\newtheoremstyle{sansrem}        
  {\topsep}{\topsep}{\normalfont}{}
  {\sffamily\itshape\color{headingcolor}}{.}{ }{}

\theoremstyle{sansplain}

\theoremstyle{sansdef}

\theoremstyle{sansrem}

\usepackage{graphicx}
\graphicspath{{figures/}}
\usepackage{subcaption}
\usepackage{booktabs}
\usepackage{multirow}
\usepackage{caption}
\usepackage[most]{tcolorbox}

\usepackage{listings}
\lstdefinestyle{mystyle}{
    backgroundcolor=\color{codebg},
    commentstyle=\color{codecomment}\itshape,
    keywordstyle=\color{codekeyword}\bfseries,
    stringstyle=\color{codestring},
    basicstyle=\ttfamily\footnotesize,
    breaklines=true, captionpos=b,
    numbers=left, numbersep=5pt, numberstyle=\tiny\color{gray},
    showstringspaces=false, tabsize=2,
    frame=single, framesep=4pt, framerule=0pt,
}
\usepackage{enumitem}
\usepackage{algorithm}
\usepackage{algpseudocode}
\usepackage{changepage}
\usepackage[textsize=footnotesize]{todonotes}

\usepackage[numbers, sort&compress]{natbib}

\usepackage{titlesec}
\titleformat{\section}
  {\sffamily\Large\bfseries\color{headingcolor}}{\thesection}{1em}{}
\titleformat{\subsection}
  {\sffamily\large\bfseries\color{headingcolor}}{\thesubsection}{1em}{}
\titleformat{\subsubsection}
  {\sffamily\normalsize\bfseries\itshape\color{headingcolor}}{\thesubsubsection}{1em}{}
\titlespacing*{\section}{0pt}{1.8\baselineskip}{0.4\baselineskip}
\titlespacing*{\subsection}{0pt}{1.0\baselineskip}{0.2\baselineskip}
\titleformat{\paragraph}[runin]
  {\sffamily\bfseries\color{headingcolor}}{}{0pt}{}
\titlespacing*{\paragraph}{0pt}{0.6\baselineskip}{0.8em}

\usepackage{hyperref}

\hypersetup{
    colorlinks=true,
    linkcolor=linkcolor,
    citecolor=linkcolor,
    urlcolor=linkcolor,
    pdftitle={Boltzmann Conditioned Molecular Generators},
    pdfauthor={Ross Irwin},
}
\usepackage[capitalize, noabbrev]{cleveref}

\newcommand{\papertitlecard}[3]{%
\begin{tcolorbox}[
enhanced, breakable,
colback=titleboxbg!10!white,
colframe=rulegray,
boxrule=0.6pt,
arc=24pt,
width=0.95\textwidth, center,
left=16pt, right=16pt, top=24pt, bottom=16pt,
before skip=0pt, after skip=28pt,
]
  \begin{center}
    {\sffamily\LARGE\bfseries\color{headingcolor} #1\par}
    \vskip 16pt
    {\sffamily\normalsize\bfseries\color{headingcolor} #2\par}
    \vskip 8pt
    {\sffamily\footnotesize\itshape\color{headingcolor} #3\par}
  \end{center}
\end{tcolorbox}
}

\newenvironment{paperabstract}{%
  \begin{center}
    {\color{rulegray}\rule{0.85\textwidth}{0.6pt}}
  \end{center}
  \vskip 12pt
  \begin{adjustwidth}{0.12\textwidth}{0.12\textwidth}
  \small\ignorespaces
}{%
  \par\end{adjustwidth}
  \vskip 4pt
  \begin{center}
    {\color{rulegray}\rule{0.85\textwidth}{0.6pt}}
  \end{center}
  \vskip 12pt
}

\begin{document}

\papertitlecard
  {Ensemble-Conditioned Molecular Design}
  {Ross Irwin\textsuperscript{1,2}$^\dagger$ \quad
   Alessandro Tibo\textsuperscript{1} \quad \\
   Jon Paul Janet\textsuperscript{1} \quad
   Simon Olsson\textsuperscript{2}}
  {%
    \begin{minipage}[t]{0.3\textwidth}
      \centering
      \textsuperscript{1}
      Molecular AI\\
      Discovery Sciences, R\&D\\
      AstraZeneca\\
      Gothenburg, Sweden
    \end{minipage}%
    \hspace{0.02\textwidth}%
    \begin{minipage}[t]{0.6\textwidth}
      \centering
      \textsuperscript{2}
      Department of Computer Science and Engineering\\
      Chalmers University of Technology\\
      and University of Gothenburg\\
      Gothenburg, Sweden
    \end{minipage}%
  }

\begingroup
\renewcommand{\thefootnote}{}
\footnotetext{$^\dagger$ Correspondence to: \texttt{rossir@chalmers.se}}
\endgroup

\begin{paperabstract}
Molecular design is typically approached as a problem of finding molecules which can adopt a single bioactive conformation. In reality, molecules occupy a distribution over conformations, and many of the properties which determine whether a candidate is viable depend on that distribution rather than on any single conformer. We reframe molecular design as an optimisation of both the modes and properties of molecules' conformational ensembles, where modes can be represented as shapes, pharmacophore profiles or protein pockets, and properties are aggregate scalars computed over the whole distribution. To realise this we introduce ensemble-conditioned guidance, a framework which conditions 3D molecular generative models on both axes simultaneously. Mode conditions are composed adaptively at inference by combining the vector fields produced under each condition. Conditions may be targeted or avoided, mixed across modalities and combined in arbitrary numbers, allowing a wide range of design tasks to be expressed with a single trained model. We introduce adaptive symmetry learning to allow conditions from different reference frames to be composed, and extend our generative framework to enable flexible-size generation. We evaluate on new benchmarks for multi-mode conditioning and ensemble property optimisation, and apply the framework to two practical drug discovery tasks, dual-target binder design and active-state-selective agonist design, where in both cases conditioning on the additional state improves the desired outcome over single-state conditioning.
\end{paperabstract}



\section{Introduction}
\label{section:introduction}

Existing methods for computational molecular design generally consider only a single conformation of a molecule. Docking~\cite{docking:canon-application} and shape-based virtual screening~\cite{shape:gaussian-vols,shape:gaussian-vol-matching,shape:rocs-application} both score a molecule by its best pose against a reference, and structure-based generative models~\cite{pocketcond:pocket2mol,pocketcond:3d-sbdd,pocketcond:target-diff,pocketcond:diff-sbdd,pocketcond:pilot,pocketcond:drug-flow,3dgen:flowr} are trained and evaluated on the bioactive conformer alone. Similarly, ligand-based generative approaches condition on the shape or pharmacophore profile of a reference conformer~\cite{shape:squid,shape:shapemol,shape:shepherd}. In reality, all molecules occupy a distribution over conformations~\cite{medchem:binding-conf-changes,medchem:binding-conf-changes-2,medchem:binding-conf-changes-3}, the Boltzmann distribution, and many crucial molecular properties are determined by that distribution rather than a single conformer.

Many practical drug design problems can be viewed through the lens of optimising both the modes and properties of a molecule's conformational ensemble. Agonist design requires a ligand which stabilises the active conformation of a receptor rather than the inactive one, and selectivity requires a molecule which binds one pocket while avoiding closely related off-targets. Dual-target design requires the opposite, a molecule which can adopt bioactive conformations in two different pockets. Beyond binding, oral bioavailability and membrane permeability depend on how much polar surface the molecule exposes across its accessible conformations~\cite{heuristics:veber,heuristics:flex-psa}, and binding affinity is closely tied to the entropic penalty flexible molecules pay on binding~\cite{heuristics:conf-entropy-binding}.

Existing approaches to these problems tend to focus on one at a time, and little attention has been paid to optimising multiple aspects of the conformational ensemble. While multi-state design has a long history in computational protein design~\cite{biogen:specific-protein-shape-design,biogen:switchable-protein-design,biogen:multistate-protein-design-framework,biogen:multistate-protein-design-review}, recent deep learning methods \cite{biogen:dynamic-mpnn,biogen:grnade,biogen:prodit,biogen:proteingenerator,biogen:caliby} have largely not carried over the ability to specify a state which should be avoided, with the exception of SwitchCraft~\cite{biogen:switch-craft}. Existing work on small molecule design tends to focus narrowly on dual-target design~\cite{multipocket:fusediff,multipocket:dualdiff}, or performs a full search over chemical space based on an oracle function~\cite{multipocket:molsculptor,multipocket:evosynth,multipocket:combimots,multipocket:multipocket-rl}. Search and optimisation approaches must be re-run for every new design, and are likely to become increasingly inefficient as further orthogonal conditions are added to the objective. Ensemble properties, meanwhile, are widely used to filter and to predict, but are rarely used to condition generation. In concurrent work, DECAF~\cite{propopt:decaf} optimises molecules towards target values of ensemble properties through iterative optimisation of a population of candidate molecules, although it does not support conditioning on specific conformational states.

\begin{figure}[t!]
    \centering
    \includegraphics[width=1.0\textwidth]{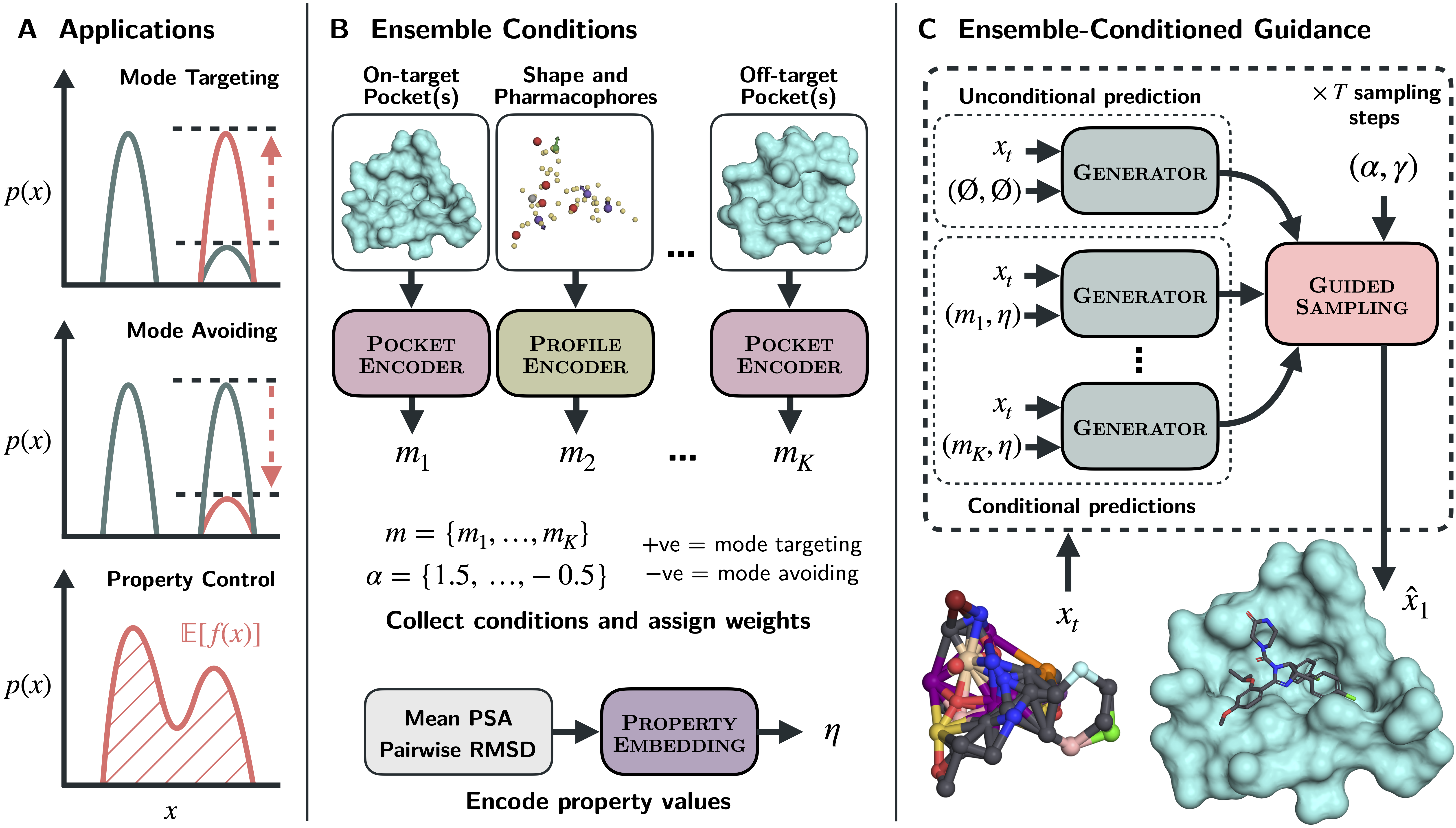}
    \caption{Overview of the \textit{ensemble-conditioned guidance} framework. \textbf{A}: Use-cases of our framework visualised in terms of how they can be used to optimise modes and properties of the conformational ensemble. \textbf{B}: Conditioning signals supported by our model; modes can be represented by shapes, pharmacophore profiles or protein pockets, and properties are target scalar values. Conditioning weights $\{ \alpha_1, \dots, \alpha_K \}$ are assigned to each mode condition, $m_i$. In practice, conditions which share a reference frame can be grouped into a single $m_i$. \textbf{C}: Guided sampling with ensemble conditions, where $x_t$ refers to the partially denoised molecule, and $\gamma$ is the strength of conditioning, analagous to classifier-free guidance.}
    \label{fig:visual-abstract}
\end{figure}

We propose to view molecular design as an optimisation over both the modes and properties of a molecule's conformational ensemble, and introduce \textit{ensemble-conditioned guidance} as a framework to realise it. Rather than training a model to satisfy several conditions jointly, we train on single conditions and compose them adaptively at inference by combining the vector fields produced under each. This side-steps the need for matched multi-mode datasets, which are scarce for any given pair of states and would have to be re-assembled for each new design task. Our framework provides considerable flexibility, unlocking a wide range of practical design tasks with a single model. Conditions can be targeted or avoided, mixed across shapes, pharmacophores, pockets and ensemble property values, and combined in arbitrary numbers. Fig.~\ref{fig:visual-abstract} provides an overview of our framework.

Composing conditions which sit in separate reference frames requires care, since combining equivariant features across frames can degrade generation. We introduce \textit{adaptive symmetry learning}, where encoders are trained to produce either invariant or equivariant features, so that one condition can define the generation frame while all others are encoded invariantly and composed without disturbing it. Separately, existing diffusion and flow models require the number of atoms to be fixed before generation. Recent work has begun to lift this restriction \cite{3dgen:flowmol3,3dgen:morph}, but not for pocket-conditioned generation, where the choice of size matters more. We introduce a simple method for generating molecules of variable size, and apply it in both single- and multi-condition settings.

Our core contributions are as follows:
\begin{itemize}
    \item \textbf{Ensemble-conditioned guidance}, allowing 3D molecular generators to be conditioned on arbitrary combinations of ensemble modes and properties at inference time.
    \item \textbf{Adaptive symmetry learning}, which allows conditioning signals to be composed without breaking the generation reference frame.
    \item \textbf{A novel encoder-decoder architecture} which encodes each condition once and supports generation of flexibly-sized molecules, keeping the cost of multi-condition sampling low.
    \item \textbf{A suite of benchmarks} for multi-mode conditioning and ensemble property optimisation, for which little prior evaluation exists.
    \item \textbf{Two case studies on practical drug discovery tasks}: dual-target binder design and active-state-selective agonist design.
\end{itemize}



\section{Background}
\label{section:background}

\subsection{Flow Matching Generative Modelling}
\label{subsection:background-fm}

\paragraph{Continuous Data}

Flow matching~\cite{fm:gaussian-fm,fm:sto-interps-iclr,fm:rectified-flows} learns a time-dependent vector field $v_\theta(x_t, t)$ that transports a simple prior $p_0$ to the data distribution $p_1$ as $t$ progresses from $0$ to $1$. Conditioning on a data endpoint $x_1 \sim p_1$ and a prior sample $x_0 \sim p_0$, training regresses $v_\theta$ against a conditional target $u_t(x_t \mid x_1)$, where $x_t$ is fixed by a chosen interpolant. Under the commonly used linear interpolant $x_t = (1-t)\, x_0 + t\, x_1$, this target reduces to $u_t = x_1 - x_0$. For molecular generative models it is common~\cite{3dgen:eqgat-diff,3dgen:semlaflow} to instead parameterise the network directly as a denoiser, predicting the endpoint $\hat{x}_1 = \hat{x}_\theta(x_t, t)$, in which case the loss takes the form
\begin{equation}
    \mathcal{L}_{\text{CFM}} = \mathbb{E}_{t, x_0, x_1} \left\| \hat{x}_\theta(x_t, t) - x_1 \right\|^2
    \label{eqn:cfm-loss}
\end{equation}
with $t \sim \mathcal{U}(0,1)$, and the velocity field is recovered at inference as $ v_{\theta}(x_t, t) = \frac{\hat{x}_1 - x_t}{1 - t} $. We adopt this endpoint parameterisation throughout. Samples are drawn by sampling a prior $x_0 \sim p_0$ and integrating the learned vector field from $t=0$ to $t=1$.

\paragraph{Discrete Flow Matching}

For categorical variables $z \in \{1, \ldots, S\}^d$ we follow~\citet{fm:multiflow}, where the generative process is a continuous-time Markov chain (CTMC) over the discrete state space. Each dimension's conditional probability path interpolates linearly on the simplex, $p_t(z \mid z_1) = (1-t)\, p_0(z) + t\, \delta_{z, z_1}$, where $\delta_{z, z_1}$ is the Kronecker delta. As above, the model is parameterised as an endpoint predictor $p_\theta(z_1 \mid z_t, t)$, trained with the cross-entropy loss
\begin{equation}
    \mathcal{L}_{\text{DFM}} = - \mathbb{E}_{t, z_0, z_1} \log p_\theta(z_1 \mid z_t, t)
    \label{eqn:dfm-loss}
\end{equation}
Since we use a uniform prior in this work, the inference-time transition rate from $z_t$ to any state $y \neq z_t$ takes the closed form $ R_t(z_t, y) = \frac{p_\theta(y \mid z_t, t)}{1 - t} $. Similarly to continuous flow matching, samples are drawn by sampling $z_0 \sim p_0$ and simulating the CTMC with Euler steps
\begin{equation}
    z_{t+\Delta t} \sim \mathrm{Cat}\bigl(\delta_{z_t,y} + R_t(z_t, y)\, \Delta t\bigr)
\end{equation}
from $t = 0$ to $t = 1$, where $y$ ranges over states. In practice, we apply additional stochasticity to this rate matrix at sample time, following~\citet{fm:multiflow}.

\subsection{Guidance for Generative Flows}
\label{subsection:background-guidance}

\paragraph{Continuous Guidance}

Inference-time conditioning for diffusion and flow models is commonly performed using guidance. Classifier guidance~\cite{guide:classifier} steers a continuous-time diffusion model towards a target conditioning $c$ by adding the gradient of the log-probability of an existing classifier $p_\phi(c \mid x_t, t)$ to the model's score estimate, with a strength parameter controlling the contribution of the classifier. Classifier-free guidance (CFG)~\cite{guide:cfg} extends this to training a single model to learn both the conditional and unconditional distributions by randomly replacing the conditioning with a null embedding $\emptyset$ during training. At inference, the conditional and unconditional outputs are linearly combined to bias the generative process. As shown in ~\citet{guide:guided-flows}, the guided velocity field of a continuous flow matching model can be written as
\begin{equation}
    v^\gamma(x_t, t, c) = (1 - \gamma)\, v_\theta(x_t, t, \varnothing) + \gamma\, v_\theta(x_t, t, c)
    \label{eqn:cfm-guide}
\end{equation}
where $\gamma$ is the guidance strength. $\gamma = 0$ recovers the unconditional flow, $\gamma = 1$ the conditional flow, and $\gamma > 1$ amplifies the conditioning signal, which has been found to improve sample quality at the cost of diversity.

Under the endpoint parameterisation used in this work, this is equivalent to combining the conditional and unconditional endpoint predictions with the same weights and recovering the guided velocity as in the unguided case, since $v_{\theta}$ is affine in $\hat{x}_{\theta}$ and the weights sum to one.

\paragraph{Discrete Guidance}

For discrete flow matching, both classifier and classifier-free guidance have been extended to act on rate-matrices~\cite{guide:discrete-fm}. In this work we focus on classifier-free guidance, where, given conditioning input $c$, and for $y \neq x_t$, the guided rate combines the conditional and unconditional rates multiplicatively:
\begin{equation}
    R_t^\gamma(z_t, y \mid c) = R_t(z_t, y \mid c)^\gamma \, R_t(z_t, y \mid \varnothing)^{1-\gamma}
    \label{eqn:dfm-guide}
\end{equation}

Importantly, under our endpoint parameterisation, the rate is proportional to the endpoint prediction, $R_t \propto p_\theta$. The multiplicative structure of Eq.~\ref{eqn:dfm-guide} therefore carries over to the endpoint predictions, $R_t^\gamma \propto p_\theta(\cdot \mid c)^\gamma p_\theta(\cdot \mid \varnothing)^{1-\gamma}$. As in the continuous case, both then take the form of a weighted combination of the model's endpoint predictions, with weights $(1-\gamma, \gamma)$ summing to one. We exploit this common structure in Section~\ref{subsection:composing-vector-fields} to extend single-condition guidance to combinations over multiple conditioning signals.


\section{Ensemble-Conditioned Guidance}
\label{section:methods}

We characterise a molecule's conformational ensemble along two axes. Ensemble modes are regions of low energy in the Boltzmann distribution, corresponding to specific 3D conformations, pharmacophore arrangements, or pocket-bound poses. Ensemble properties are aggregate scalars computed over the distribution, such as mean polar surface area or mean pairwise RMSD. Here, we introduce \textit{ensemble-conditioned guidance} to condition on both axes simultaneously, and compose signals for multiple ensemble modes at inference time. A single trained model can therefore target one or more modes while controlling a particular ensemble property, even when the objectives pull in different directions.

Crucially, unlike previous related methods for multi-state design~\cite{biogen:grnade,biogen:dynamic-mpnn,multipocket:fusediff}, this composition does not require matched multi-mode conditions at training time, allowing us to use regular molecular and protein-ligand datasets without constraints. Below we outline how ensemble conditions are represented, then describe our molecular representation and model architecture, our method for dynamic vector field composition, and finally our training setup.


\subsection{Ensemble Conditions}
\label{subsection:ensemble-conditions}

Modes can be encoded using explicit shape and pharmacophore spatial conditions, or through a \textit{soft} protein pocket constraint, where the model is given freedom to generate the binding conformation. Ensemble properties are encoded directly by target value, as described further below.

\paragraph{Shape Conditioning}

Taking inspiration from Gaussian volume overlap methods for molecular shape matching~\cite{shape:gaussian-vols,shape:gaussian-vol-matching}, we represent a conformer's shape as a noisy version of its coordinates. We first duplicate each atomic coordinate independently with probability $0.2$ (to ensure the model cannot infer the exact atom count of the reference conformer) and then convolve each resulting point with isotropic Gaussian noise drawn from $\mathcal{N} (0, \sigma_{shape}^2\mathbf{I})$. The subsequent noisy coordinates are then used to condition the generative model on the given shape, where $\sigma_{shape}$ controls the fidelity of the condition. During training we sample $\sigma_{shape} \sim \mathcal{U} (0.1, 1.0)$, thereby allowing a choice of conditioning fidelity at inference.

\paragraph{Pharmacophore Conditioning}

Pharmacophores (as well as reference protein-ligand interactions) are similarly represented using 3D points. Our pharmacophore conditioning supports: hydrogen bond donors and acceptors; cations and anions; aromatic rings; and hydrophobic regions. For protein-ligand data, key binding interactions are mapped to one of these groups (full details in Appendix~\ref{appendix:extended-methods}). As in ShEPhERD~\cite{shape:shepherd}, we include a direction vector for pharmacophores, but only for hydrogen bond donors (direction from the heavy atom to its connected hydrogen) and aromatic rings (the normal to the ring plane); all others are set to a zero vector. Each pharmacophore is therefore represented as a 3D coordinate, a direction vector and an interaction type. We randomly drop out pharmacophores during training, with a dropout rate of 0.5 applied independently. A dropout probability of 0.2 is used for protein-ligand data, where crystal interactions are known.

\paragraph{Pocket Conditioning}

As well as \textit{hard} mode constraints, where the model is asked to match a specific conformer shape or pharmacophore point cloud, our method supports \textit{soft} mode conditioning based on protein pockets. Soft conditions ask the model to generate a molecule which binds to the given site but give the model freedom to generate its chosen binding conformation. The model is provided with the atom types and coordinates of the pocket residues -- extracted in advance at a 6Å radius from a reference ligand.

\paragraph{Property Conditioning}

Our method also supports generation conditioned on properties computed over the conformational ensemble. While our framework in principle supports conditioning on any function over the pre-computed ensemble, we focus on the following commonly-used heuristics:
\begin{itemize}
    \item Polar surface area (PSA) measures the 3D surface area of polar atoms (we use nitrogen, oxygen and any hydrogen bonded to a nitrogen or oxygen). PSA is frequently used as a heuristic to guide orally-bioavailable and membrane-permeable drug design~\cite{heuristics:veber,heuristics:flex-psa}. We compute it using RDKit's implementation of FreeSASA for each conformer. For the remainder of this paper we use PSA to refer to the mean PSA, averaged over the ensemble.
    \item Mean pairwise root-mean-squared deviation (RMSD) quantifies the flexibility of the molecule by measuring the average RMSD between all pairs of conformers after superpositioning. Controlling flexibility is crucial for designing potent drugs since flexible molecules pay a higher entropic cost when binding to a target, reducing affinity. In practice, we cap the number of pairs at 200 to avoid a runtime explosion.
\end{itemize}

To maximise inference-time flexibility, we randomly mask components of the input conditions during training. Shape profiles, pharmacophore profiles and each property value are dropped independently, so the model sees many combinations of conditioning signals and remains a valid predictor for any subset chosen at inference. To create the null condition $\varnothing$ for classifier-free guidance, we additionally drop all signals jointly, including protein pockets.


\subsection{Molecular Representation and Generative Process}
\label{subsection:fm-training-setup}

We now outline the molecular representation used by our framework. We follow a similar setup to the one described in~\citet{3dgen:semlaflow}. Notably, however, we extend their framework to enable generation of arbitrarily-sized molecules, as described below.

\begin{figure}[t]
    \centering
    \includegraphics[width=\textwidth]{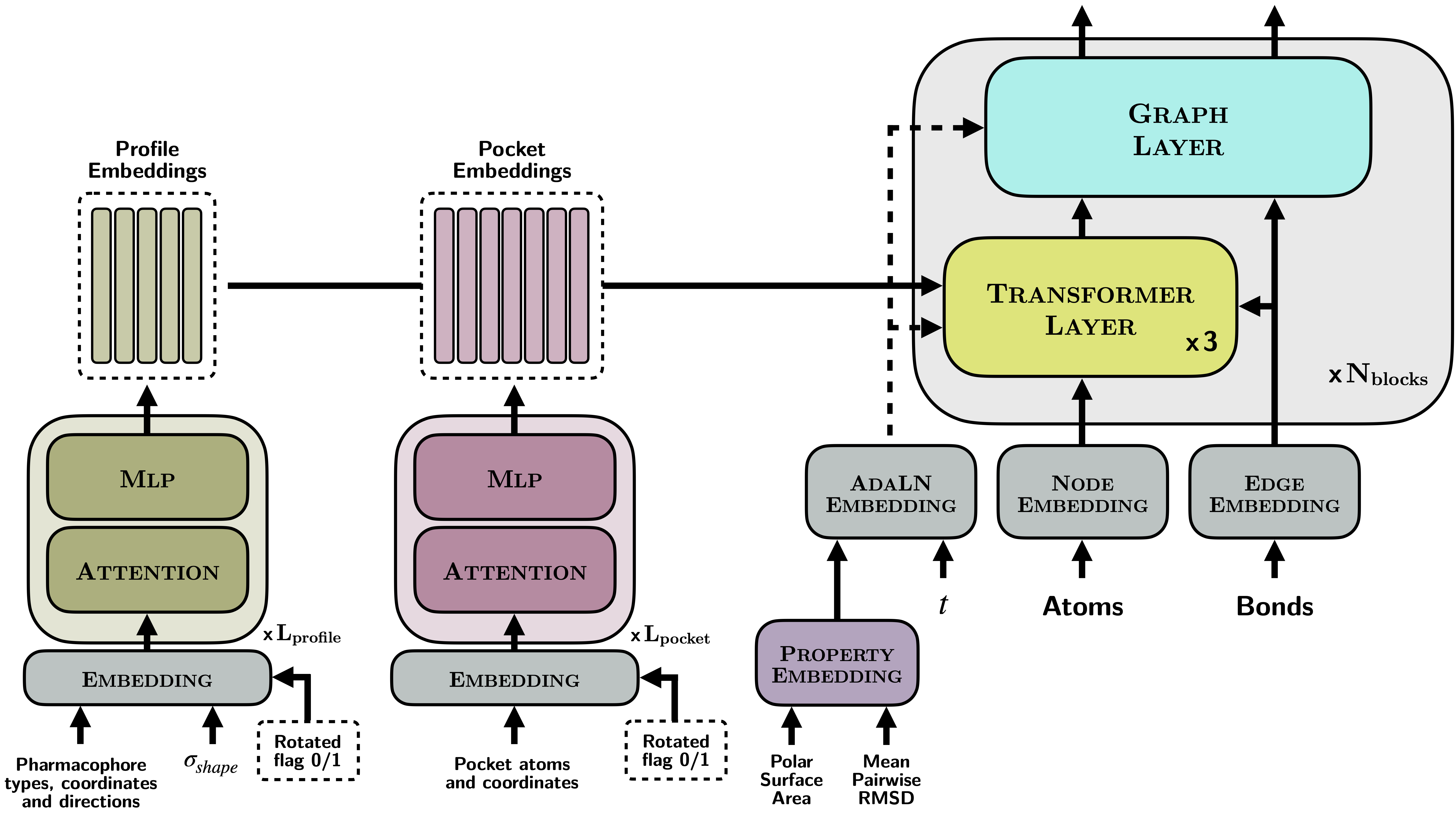}
    \caption{Our model follows an encoder-decoder design, with separate encoders for shape/pharmacophore profiles and protein pockets, as well as an ensemble property embedding module. Profile and pocket encoders also make use of adaptive layer normalisation (not shown for brevity). Encoders follow a standard transformer design, while the decoder mixes transformer with graph transformer layers, along with adaptive layer normalisation conditioning for incorporating property conditions and the flow-matching time $t$.}
    \label{fig:model-overview}
\end{figure}

\paragraph{Molecule Representation}

A molecule is represented as a tuple $(X, a, b)$ of heavy-atom coordinates $X \in \mathbb{R}^{N \times 3}$, atom types $a \in \{1, \dots, S_{a}\}^N$, and a pairwise bond matrix $b \in \{1, \dots, S_b\}^{N \times N}$. Each atom's element and formal charge are bundled into a single categorical token, with a dedicated padding token used for the fixed-size representation that enables flexible-size generation (see below). Bonds are categorical over bond order and aromaticity, with a \textit{no-bond} category for non-bonded atom pairs. Hydrogen atoms are not modelled explicitly, but instead added after the sampled molecule is converted to RDKit.

\paragraph{Joint Continuous-Discrete Flow Matching}

We treat the tuple $x = (X, a, b)$ as a single data sample, modelling coordinates with continuous flow matching (Eq.~\ref{eqn:cfm-loss}) and atom and bond types with discrete flow matching (Eq.~\ref{eqn:dfm-loss}). The prior $p_0$ factorises across modalities: we use a standard Gaussian with zero centre-of-mass for coordinates, and independent uniform categoricals for each atom and bond token. Training samples $x_t$ are constructed by drawing $x_0 \sim p_0$, $x_1 \sim p_1$, and sampling a time $t \in [0,1]$. Coordinates are interpolated using the Gaussian-perturbed linear path from~\citet{fm:ot-cfm}, $X_t \sim \mathcal{N} ((1 - t)X_0 + tX_1 , \sigma_x^2 \mathbf{I})$, where $\sigma_x = 0.2$\,\AA. Atom and bond types are interpolated independently per token using the linear categorical path discussed in Section~\ref{subsection:background-fm}.

We parameterise the network as a denoiser which predicts the endpoint
$\hat{x}_1 = (\hat{X}_1, \hat{a}_1, \hat{b}_1)$, where $\hat{X}_1 \in
\mathbb{R}^{N \times 3}$ is a coordinate estimate, and $\hat{a}_1$
and $\hat{b}_1$ collect categorical distributions over atoms and
bonds, respectively, for each token. Combinations of endpoint predictions therefore act linearly on the coordinates and log-linearly on the categorical distributions, which we make use of in Section~\ref{subsection:composing-vector-fields}. At inference, molecules are generated by sampling $x_0 \sim p_0$ and integrating the learned velocity field from $t = 0$ to $t = 1$ with Euler steps using the procedure outlined in Section~\ref{subsection:background-fm}. Following~\citet{3dgen:semlaflow}, we use a geometric decay schedule during sampling (smaller steps as $t \to 1$), which has been found to improve sample quality. We use $100$ sampling steps unless stated otherwise.

\paragraph{Flexible-Size Generation}

Most existing diffusion and flow models for molecular generation require the number of atoms in a molecule to be fixed in advance of generation. To support flexible-size generation we propose to give the model a fixed number of atom tokens; padding all molecules to a fixed maximum size (we use 48 atoms) and training the model to generate molecules including padding atoms. Sampling simply starts from a fixed-size set of random atoms and the model assigns padding tokens to unused atoms. We found two elements of the training setup to be critical for learning the correct size distribution (from unconditional sampling):
\begin{enumerate}
    \item Setting coordinates for all \textit{pad} atoms as the centre-of-mass of the molecule.
    \item Applying a permutation alignment between the padded versions of $x_0$ and $x_1$, which minimises the total squared distance between paired prior and data atoms (after centring both point clouds at the origin so that translations don't dominate the cost).
\end{enumerate}
Together these can be seen as an extension of equivariant optimal transport paths~\cite{confgen:equi-ot,3dgen:equifm} to flexible-size generation. We hypothesise these are useful for learning the correct size distribution as they force the model to place pad atoms close to the centre of mass of the molecule, potentially improving the training signal. We study these decisions further in Appendix~\ref{appendix:size-learning}.


\subsection{Architecture}
\label{subsection:arch}

At a high-level, our model resembles an encoder-decoder neural network architecture, where ensemble conditions (modes and scalar ensemble properties) are encoded into a stack of vectors. These are then passed to the decoder which is trained to generate molecules satisfying the given conditions. We use adaptive layer normalisation (AdaLN)~\cite{arch:adaln} throughout, with zero-initialised output scaling parameters, following~\citet{arch:dit}. Fig.~\ref{fig:model-overview} outlines our architecture.

\paragraph{Encoding Conditions}

Shape and pharmacophore points are embedded jointly by a shared encoder; we refer to this set as a \textit{profile}. Protein pockets are embedded separately. Both encoders are lightweight, non-equivariant transformers, embedding geometric information jointly with invariant node features. Property values are projected through a learnable MLP and masked-summed into a single conditioning vector, where dropped properties contribute nothing to the sum.

\paragraph{Generating Molecules}

We denote the generator as $\hat x_\theta(x_t, t, m, \eta)$, where $m$ collects the cross-attention conditioning signals (profile and pocket) and $\eta$ refers to the property values fed through AdaLN. These together play the role of the single conditioning input $c$ in Section~\ref{subsection:background-guidance}. Either input can be replaced by a null token $\varnothing$, in which case the corresponding signal is dropped. The decoder is a stack of blocks, each mixing transformer layers with graph transformer layers in a 3:1 ratio, balancing efficiency with expressivity while supporting both conditional generation and pairwise bond prediction. Transformer layers cross-attend to the conditioning embeddings $m$; graph transformer layers operate purely on the partially denoised graph and are used to update pairwise features. Further details on our transformer and graph transformer layers, including our graph latent attention mechanism adapted from SemlaFlow~\cite{3dgen:semlaflow}, are provided in Appendix~\ref{appendix:arch-details}.

\paragraph{Adaptive Symmetry Learning}

Since we wish to sample using potentially many structural conditions, we require a robust mechanism for combining conditioning signals during inference. In early experiments, we found that directly combining equivariant conditioning features from different reference frames led to degraded performance. Inspired by work on learning molecular symmetries through augmentation~\cite{confgen:mcf,mlip:escaip,3dgen:adit}, we propose to train our encoders to embed \textit{adaptive symmetries} -- where we can select whether encoded features contain E(3)-invariant or -equivariant signal. To achieve this, we randomly rotate encoder inputs out of their reference frame 50\% of the time, and pass the binary \textit{rotated} flag to the encoder. We expect the model to treat the signal as invariant when a rotation has been applied, and equivariant when it has not. In Appendix~\ref{appendix:adaptive-symmetry} we study how well the model learns these symmetries. In practice, we find that the symmetry error varies significantly over the generation trajectory, but all errors converge towards zero at $t=1$. This suggests that, since generation is built on iterative denoising, encoding exact symmetries into the model may be unimportant.


\subsection{Composing Vector Fields for Multi-Mode Conditioning}
\label{subsection:composing-vector-fields}

To construct composite signals that target multiple modes simultaneously, or push generation away from undesired modes, we compose mode signals adaptively at inference. In this section we outline our method for multi-mode conditioning.

\paragraph{Multi-Mode Guidance}

The combinations of Eqs.~\ref{eqn:cfm-guide} and~\ref{eqn:dfm-guide} act independently on each output of the generator: the continuous combination on the coordinate endpoint, and the discrete log-linear combination on each categorical distribution:
\begin{equation}
    \hat{X}^\gamma = (1-\gamma)\, \hat{X}_\theta(x_t, t, \varnothing, \varnothing) + \gamma\, \hat{X}_\theta(x_t, t, m, \eta)
    \label{eqn:cfg-continuous} 
\end{equation}
\begin{equation}
    \log p^\gamma(z_1 \mid x_t, t) = (1-\gamma) \log p_\theta(z_1 \mid x_t, t, \varnothing, \varnothing) + \gamma \log p_\theta(z_1 \mid x_t, t, m, \eta), \quad z \in \{ a, b\}
    \label{eqn:cfg-discrete}
\end{equation}
Both forms share weights $(1-\gamma, \gamma)$ summing to one. For simplicity, we use the same weights across coordinates, atoms and bonds throughout this paper.

This structure generalises to composition over an arbitrary number of conditioning signals. Given conditions $m_1, \dots, m_K$ and per-condition strengths $\gamma_1, \dots, \gamma_K$, we define the composed endpoint predictions as:
\begin{equation}
    \hat{X}^{\,\gamma}
    = \Big(1 - \sum_{k=1}^{K} \gamma_k\Big)\,
      \hat{X}_\theta(x_t, t, \varnothing, \varnothing)
    + \sum_{k=1}^{K} \gamma_k\, \hat{X}_\theta(x_t, t, m_k, \eta)
    \label{eqn:compose-continuous}
\end{equation}
\begin{equation}
    \log p^{\gamma}(z_1 \mid x_t, t)
    = \Big(1 - \sum_{k=1}^{K} \gamma_k\Big)
      \log p_\theta(z_1 \mid x_t, t, \varnothing, \varnothing)
    + \sum_{k=1}^{K} \gamma_k \log p_\theta(z_1 \mid x_t, t, m_k, \eta),
    \quad z \in \{a, b\}
    \label{eqn:compose-discrete}
\end{equation}
where $\gamma = (\gamma_1, \dots, \gamma_K)$ and the null prediction absorbs the residual weight, so that the full set of weights sums to one and the endpoint-velocity equivalence of Section~\ref{subsection:background-guidance} applies unchanged. Each $\gamma_k$ therefore keeps its single-condition meaning, the strength with which $m_k$ is applied, with $\gamma_k < 0$ pushing generation away from it, and $K=1$ recovers Eqs.~\ref{eqn:cfg-continuous} and \ref{eqn:cfg-discrete}. More generally, Eq.~\ref{eqn:compose-continuous} composes vector fields as in compositional diffusion~\cite{multimode:comp-gen-energy-models,multimode:comp-diff} and Eq.~\ref{eqn:compose-discrete} is a weighted product of experts~\cite{multimode:p-of-e}, with negative $\gamma_k$ acting as the negation operator.

It is often convenient to separate how far the composition extrapolates from the unconditional flow from how that extrapolation is divided between conditions. Writing $\gamma_k = \gamma \alpha_k$ with $\sum_{k} \alpha_k = 1$, the scalar $\gamma$ recovers its usual role as the overall guidance strength while $\alpha$ specifies the allocation across conditions, with $\alpha_k < 0$ for an avoided condition. We report all experiments in this parameterisation.

\paragraph{Reference Frame-Guided Generation}

Since it is often desirable to support frame-aware generation (generating a protein-bound conformation, for example), we select one $m_k$ as the reference for generation and toggle the encoder to produce equivariant features using the adaptive symmetry mechanism of Section~\ref{subsection:arch}. All other mode conditions are encoded invariantly, which permits their composition without breaking the reference frame. In practice, for generation efficiency, a single condition $m_k$ may itself bundle multiple embeddings, provided they share a reference frame. For example, profile (shape and pharmacophore) embeddings and pocket embeddings derived from the same ligand can be concatenated as a single $m_k$, requiring only one forward pass through the generator.

\paragraph{Practical Application}

Because composition is constructed entirely at inference, the model only sees
single-mode samples during training. The independent condition masking described
in Section~\ref{subsection:ensemble-conditions} ensures that each
$\hat{x}_\theta(\cdot, \cdot, m_k, \eta)$ remains valid under whichever subset of
conditions is chosen at test time. This form of dynamic signal composition also
permits significant inference-time flexibility. We break conditioning strategies down into the following three regimes of practical interest:
\begin{enumerate}
    \item \textbf{Mode targeting} With all $\alpha_k > 0$, generation is steered
    toward a region jointly satisfying all $K$ modes.
    \item \textbf{Mode avoidance} With $\alpha_k < 0$ for some $k$, generation is
    pushed away from $m_k$.
    \item \textbf{Signal amplification} With $\gamma > 1$, the combination
    extrapolates beyond the conditional flow, in direct analogy to standard CFG
    amplification. This is independent of the choice of $\alpha$ and can be
    combined with either regime above.
\end{enumerate}
These regimes can also be combined within a single composition; for example, targeting a protein pocket with a set of reference pharmacophores while concurrently avoiding two off-target pockets. Algorithm~\ref{algo:sampling} summarises our general sampling procedure.

\begin{algorithm}[t!]
    \caption{Sampling using ensemble-conditioned guidance.}
    \label{algo:sampling}
    \begin{algorithmic}[1]
        \Require Prior $x_0$, time schedule $\{t_0, \dots, t_T\}$, embedded mode conditions $\{m_1, \dots, m_K\}$, property embedding $\eta$, per-condition strengths $\gamma_k = \gamma \alpha_k$ with $\sum_k \alpha_k = 1$
        \State $x \gets x_0$
        \For{$i = 0, \dots, T-1$}
            \State $\hat x^{(0)} \gets \hat x_\theta(x, t_i, \varnothing, \varnothing)$  \Comment{null prediction}
            \For{$k = 1, \dots, K$}
                \State $\hat x^{(k)} \gets \hat x_\theta(x, t_i, m_k, \eta)$  \Comment{conditional prediction}
            \EndFor
            \State Combine endpoint predictions using Eqs.~\eqref{eqn:compose-continuous} and~\eqref{eqn:compose-discrete} with strengths $\{\gamma_1, \dots , \gamma_K\}$
            \State Integrate from $t_i$ to $t_{i+1}$ using the procedure from Section~\ref{subsection:background-fm}
        \EndFor
        \State \Return $x$
    \end{algorithmic}
\end{algorithm}


\subsection{Model Training}
\label{subsection:model-training}

As outlined in Section~\ref{subsection:fm-training-setup}, we train the network to predict the endpoint $\hat{x}_1 = (\hat{X}_1, \hat{a}_1, \hat{b}_1)$ from the interpolated sample $x_t$. We train with the combined objective:
\begin{equation}
    \mathcal{L} = \mathbb{E}_{t,x_0,x_1} \left[ \omega(t) \cdot \left( \| \hat{X}_1 - X_1 \|^2 + \lambda_a \, \mathrm{CE}(\hat{a}_1, a_1) + \lambda_b \, \mathrm{CE}(\hat{b}_1, b_1) \right) \right],
    \label{eqn:joint-loss}
\end{equation}
where $\mathrm{CE}$ is the cross-entropy and $\omega(t) = \min \big( \frac{t}{1-t}, 10.0 \big)$. We use $\lambda_a = 0.3$ and $\lambda_b = 5.0$, chosen to balance the losses for the discrete and continuous modalities.

Training samples are constructed by drawing $x_0 \sim p_0$, $x_1 \sim p_1$ and $t \sim \mathrm{Beta}(1.5, 1.0)$ rather than the uniform time sampling of Eq.~\ref{eqn:cfm-loss}. Together with $\omega$, this biases training towards late $t$, following~\cite{3dgen:semlaflow,3dgen:tabasco}. Conditions are masked during training as described in Section~\ref{subsection:ensemble-conditions}. A masking rate of 30\% is applied independently to the shape profile, pharmacophore profile, PSA and pairwise RMSD, and a global rate of 10\% is applied to all signals including protein pockets.

The model is trained jointly on two datasets: GEOM Drugs~\cite{datasets:geom}, consisting of 300K small molecules with conformer ensembles pre-computed using CREST~\cite{ff:crest}, and SPINDR~\cite{3dgen:flowr}, a dataset of 35K protein-ligand complexes originally derived from the PDB, with pre-computed protein-ligand interactions. For GEOM Drugs, we hold out a test set of 1K molecules consisting of novel, unique scaffolds (Appendix~\ref{appendix:data-splits}), which we use to construct our multi-mode benchmark (Section~\ref{subsection:multi-cond-expts}). SPINDR splits are the same as those proposed by PLINDER~\cite{datasets:plinder}, which were chosen to minimise train-test data leakage. During training, we filter out SPINDR systems with QED $< 0.3$ and replicate the remaining 28K SPINDR systems 8 times each epoch. Training was performed in mixed-precision (bf16) using the Adam optimiser for 200 epochs on a single A100 GPU, taking approximately 2 days. Full hyperparameters are listed in Appendix~\ref{appendix:arch-details}.


\section{Experiments}
\label{section:results}

Since there is limited existing work on designing molecules which adopt specified conformational states or target ensemble properties, we set up a suite of challenging benchmarks to test our method. We investigate both multi-mode generation (mode targeting and mode avoidance), and optimising an ensemble property while maintaining a bioactive pharmacophore profile. In Section~\ref{subsection:case-studies} we apply our model to two real-world drug discovery tasks -- dual-target binder design, and agonist design (optimising for active-state selectivity).

To keep our evaluations tractable, we set up a conformer sampling algorithm using ETKDG~\cite{ff:etkdg} (implemented in RDKit), and minimisation with the MMFF94 forcefield~\cite{ff:mmff}. This bypasses the use of CREST~\cite{ff:crest}, which typically takes hours per molecule, and we find it provides a reasonable approximation for our ensemble properties (full details in Appendix~\ref{appendix:ensemble-sampling}).


\subsection{Multi-Mode Conditioning}
\label{subsection:multi-cond-expts}

\begin{figure}[t!]
    \centering
    \includegraphics[width=\textwidth]{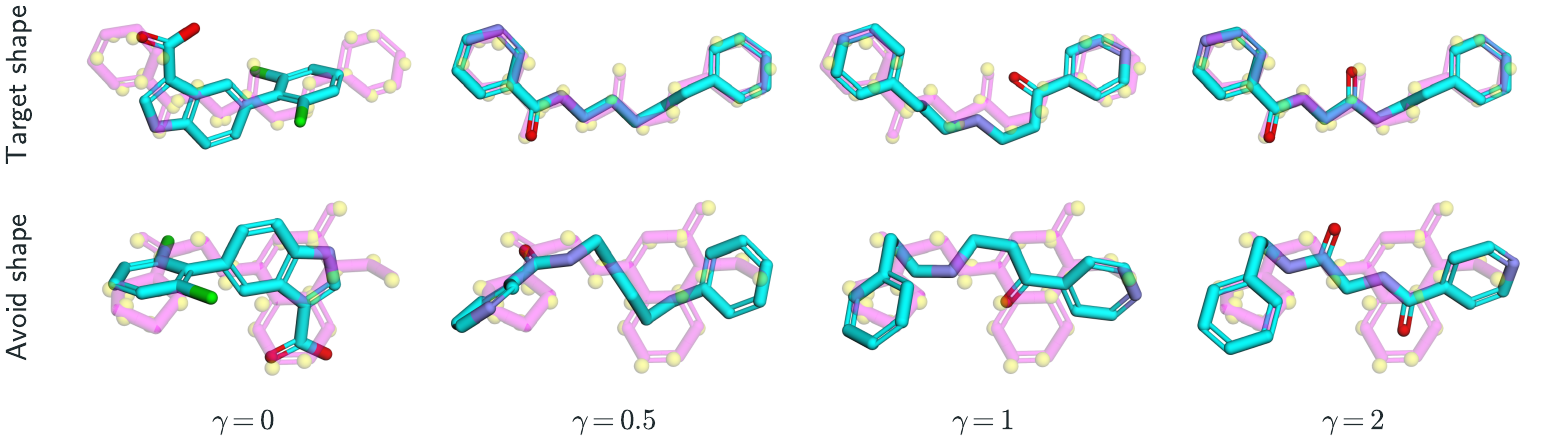}
    \caption{Generating molecules to target an extended shape while avoiding a compact shape for different conditioning strengths, $\gamma$, where $\gamma = 0$ refers to unconditional generation. The two reference molecules are shown in magenta, while yellow points show the shape profiles used for conditioning. The conformer with the highest Gaussian shape overlap to the reference is plotted, and $\sigma_{shape} = 0.2$ was used in all cases.}
    \label{fig:mode-avoid-cfg-grid}
\end{figure}

These benchmarks aim to evaluate mode targeting and mode avoidance abilities. In order to test fundamental capabilities, we restrict ourselves to conditioning on only two modes, which are represented using shape profiles only (Section~\ref{subsection:ensemble-conditions}). Fig.~\ref{fig:mode-avoid-cfg-grid} shows an example of molecules generated to target an extended shape, while avoiding a compact shape, for different values of $\gamma$.

\paragraph{Experimental Setup}

We sample MMFF ensembles for all molecules in our GEOM Drugs test set (1K molecules) using the procedure above, and select a compact and an extended conformer for each based on the radius of gyration. We then find two complementary sets of compact-extended pairings: mode targeting pairings, where molecules fit both modes; and mode avoiding pairings, where molecules fit one mode but not the other. To create meaningful pairings, we ensure there exists at least one chemically-distinct reference molecule in the training set which fits both mode requirements, and only pair conformers whose parent molecules differ by at most one heavy atom. The mode targeting set contains 496 pairs, while the mode avoiding set contains 535. Full details of the benchmark construction are given in Appendix~\ref{appendix:multi-cond-benchmark}.

Each molecule is scored against a target shape by sampling its MMFF ensemble, scoring every conformer against the target via Gaussian shape overlap (using RDKit \texttt{rdShapeAlign}), and taking the best conformer's shape Tanimoto. For each target we define a size-matched virtual screening baseline by taking the mean of this score over a random sample of training molecules within 1 heavy atom of the reference. We report the shape-matching gain $\Delta$ as the difference between the generated tanimoto scores and the baseline for the compact and extended targets separately. $\Delta$ therefore measures how much better a generated molecule fits the target shape than a randomly drawn molecule of the same size.

\paragraph{Results}

\begin{figure}[t!]
    \centering
    \includegraphics[width=\textwidth]{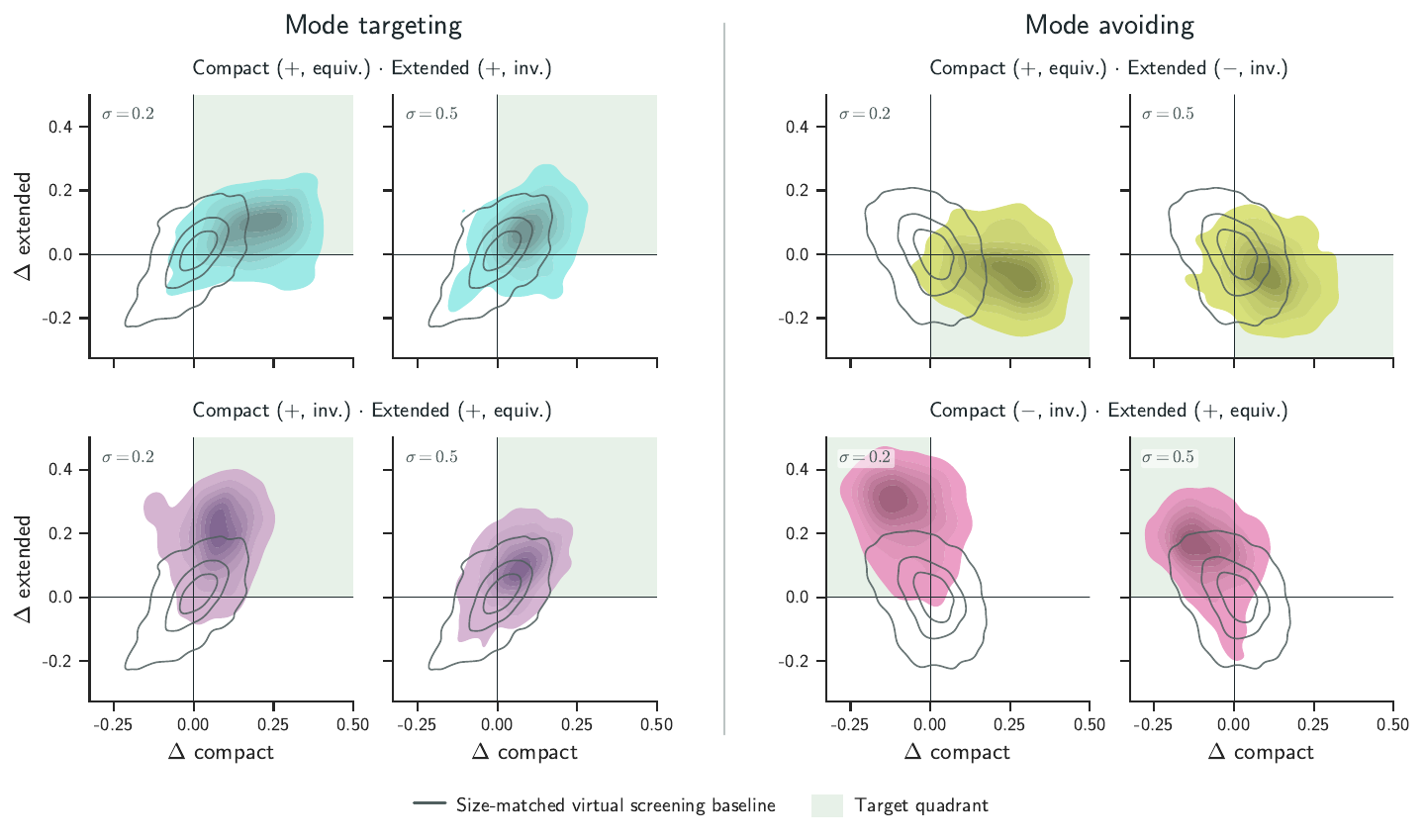}
    \caption{Density plots showing the difference in shape Tanimoto scores compared to a size-matched baseline on mode targeting and mode avoiding tasks. $+$ and $-$ correspond to mode targeting and mode avoiding conditioning, respectively, and equiv. and inv. correspond to using equivariant and invariant conditioning signals.}
    \label{fig:multi-cond-delta}
\end{figure}

In Fig.~\ref{fig:multi-cond-delta} we show the density plots of compact and extended $\Delta$ when running our model on our mode targeting and avoiding benchmarks, using values of 0.2 and 0.5 for $\sigma_{shape}$. In all runs $\gamma = 4.0$, with $\alpha = (0.5, 0.5) $ for mode targeting and $\alpha = (1.5, -0.5)$ for mode avoidance, where the first entry refers to the targeted mode and the second to the avoided one. In general we see a strong shift away from the size-matched virtual screening baseline and toward the target condition, which is particularly prominent at lower $\sigma_{shape}$. This highlights the useful role of $\sigma_{shape}$ as controlling the conditioning fidelity; as more noise is added to the shape profile, the generated molecule fits the target mode less strongly, although we find this leads to samples which are more chemically distinct from the reference (see Appendix~\ref{appendix:multi-shape-cond-results}). Additionally, the plots also show a bias in conditioning strength towards the condition that was provided with equivariant features, especially for $\sigma_{shape} = 0.2$. When targeting both modes (left hand side of Fig.~\ref{fig:multi-cond-delta}), the dominant axis of response shifts when the mode given equivariant features is flipped. At $\sigma_{shape} = 0.2$ the mean compact and extended $\Delta$ are $0.19$ and $0.07$ when the compact mode is given equivariant features, and $0.07$ and $0.19$ when the assignment is reversed, showing that the bias follows the feature type rather than the mode.

Importantly, we also find that generation quality and chemical novelty are maintained across all four conditioning setups. Connected validity stays above $0.97$ and uniqueness above $0.99$ in every run except compact avoidance at $\sigma_{shape} = 0.2$, where it falls to $0.90$. Generated molecules also remain chemically distinct from the two reference molecules used to build the conditions, with a mean ECFP Tanimoto of between $0.25$ and $0.31$ at $\sigma_{shape} = 0.2$ and $0.14$ at $\sigma_{shape} = 0.5$ in all four setups. Taking the more similar of the two references gives medians of $0.37$ and $0.15$ respectively, so as well as controlling how strongly the target shape is matched, $\sigma_{shape}$ trades conditioning fidelity against chemical novelty. Full per-run results are given in Table~\ref{tab:full-multicond-results}.

\subsection{Ensemble Property Optimisation}
\label{subsection:prop-opt-expts}

This benchmark aims to evaluate whether ensemble properties can be controlled while a bioactive conformation is maintained. We condition on a protein pocket together with the reference ligand's interactions, which fixes the binding mode, and additionally ask for a target value of a single ensemble property, either mean PSA or mean pairwise RMSD (Section~\ref{subsection:ensemble-conditions}). Since one condition constrains the bound conformation while the other constrains the whole ensemble, we measure both the property control achieved and the cost paid in pocket fit.

\paragraph{Experimental Setup}

We take protein-ligand systems from the test set of SPINDR, which are split to be structurally distinct from the training systems~\cite{datasets:plinder}. Following the same filter used during training, we remove any system whose reference ligand has a QED below $0.3$, leaving 179 test systems, and generate 3 molecules for each. All runs condition on the protein pocket together with the pharmacophores extracted from the reference ligand's crystal interactions, using $\gamma = 2.0$. Since the pocket and its reference pharmacophores share a reference frame, they are encoded as a single mode, so $K = 1$ and $\alpha = (1)$. Property-conditioned runs additionally provide a target for one ensemble property, where the value given to the model is drawn from $\mathcal{N}(\mu_{target}, \sigma_{prop})$, with $\sigma_{prop}$ of 10 Å$^2$ for PSA and 0.1 Å for mean pairwise RMSD. We sweep PSA targets of 80, 100, 120 and 140 Å$^2$, and mean pairwise RMSD targets of 1.0, 1.5, 2.0 and 2.5 Å, and compare against a baseline using the same pocket and pharmacophore conditioning with no property condition. 

Generated molecules are scored for ensemble properties using the sampling procedure above, for the fraction of reference protein-ligand interactions recovered, and by scoring each generated pose in its pocket with AutoDock Vina~\cite{docking:vina}, both as generated and after local minimisation under the Vina scoring function. Interaction recovery measures the proportion of conditioned reference interactions for which the generated molecule places a pharmacophore of the same type within 2Å. Since a model could satisfy this using a highly strained conformer, we first apply a short local GFN2-xTB relaxation on the generated pose. Full definitions of all evaluation metrics are given in Appendix~\ref{appendix:eval-metrics}.

\begin{figure}[t!]
    \centering
    \hfill
    \begin{subfigure}[b]{0.49\textwidth}
        \centering
        \includegraphics[width=\textwidth]{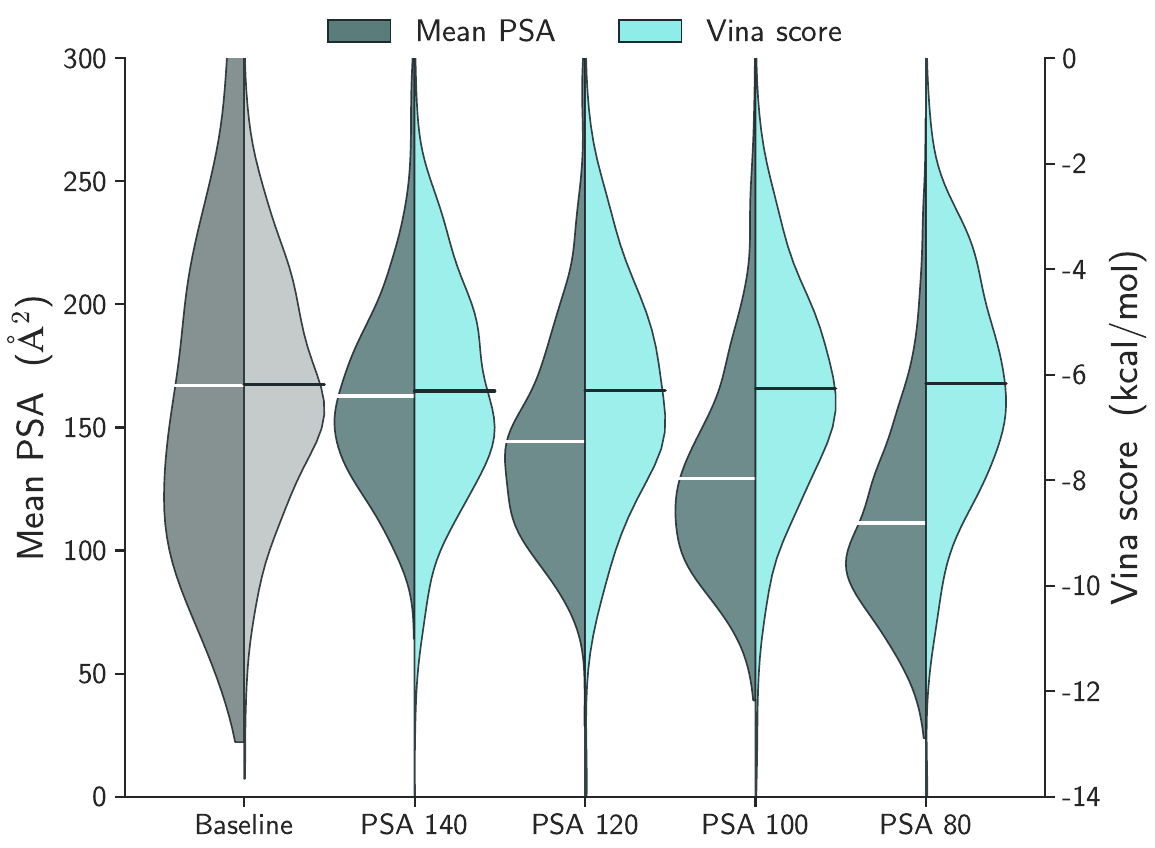}
    \end{subfigure}
    \hfill
    \begin{subfigure}[b]{0.49\textwidth}
        \centering
        \includegraphics[width=\textwidth]{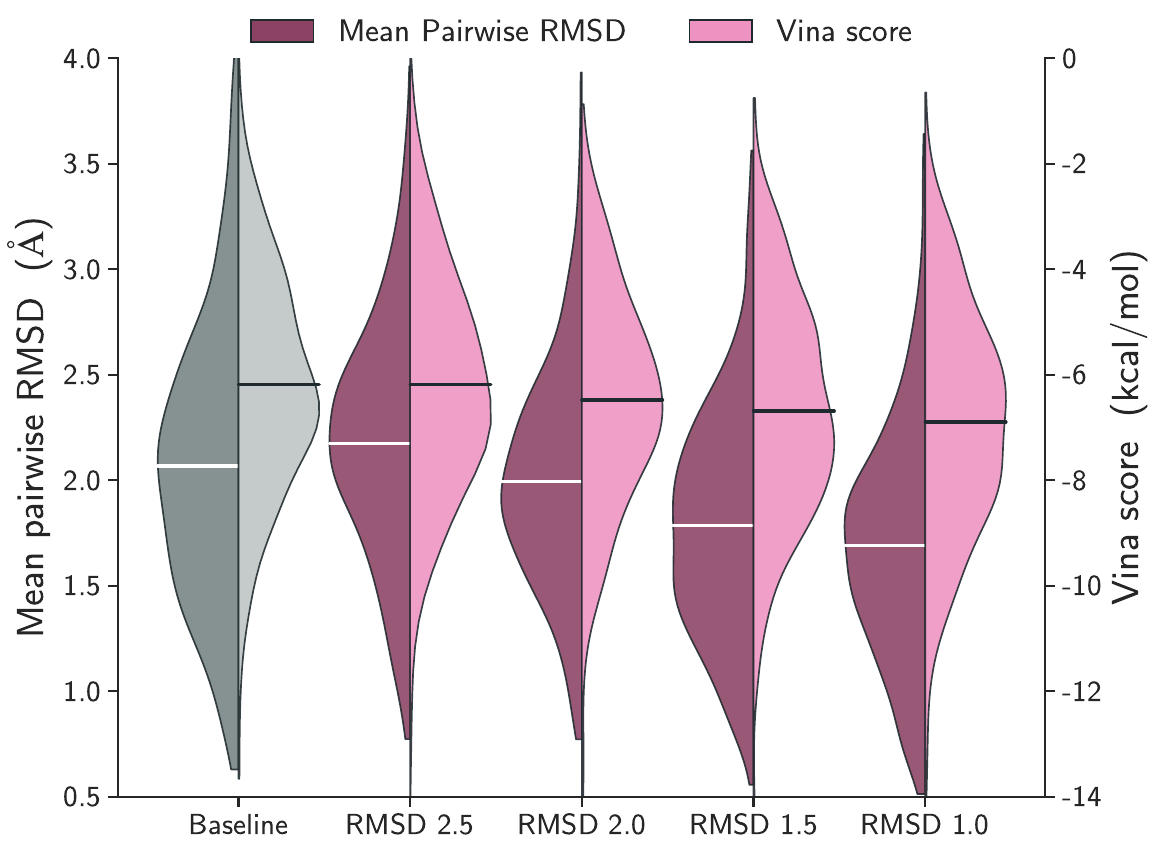}
    \end{subfigure}
    \caption{Distribution plots for molecular properties generated from the ensemble property optimisation benchmark. \textit{Baseline}: pocket and pharmacophore conditioning; other plots come from sampling with the same condition as well as a property condition. Centre lines denote the mean of the distributions.}
    \label{fig:prop-opt-violins}
\end{figure}

\paragraph{Results}

As shown in Fig.~\ref{fig:prop-opt-violins}, we find that both properties respond monotonically to the conditioning target. Mean PSA moves from 111 to 163 Å$^2$ as the target is raised from 80 to 140, and mean pairwise RMSD from 1.69 to 2.17 Å as the target is raised from 1.0 to 2.5, against baseline values of 167 Å$^2$ and 2.07 Å. Achieved values are compressed towards the baseline at the ends of each sweep, most visibly for PSA, where a target of 80 produces a mean of 111 Å$^2$. The model therefore provides directional control over both properties across a wide range, rather than calibrated control at a specific value.

Crucially, this control does not come at the cost of the pocket conditioning. Interaction recovery stays between $0.944$ and $0.961$ for every run, matching the $0.954$ of the pocket and pharmacophore baseline, and Vina scores, validity and similarity to the reference ligand are all maintained across both sweeps. Docking scores even improve slightly at lower RMSD targets, from the baseline's $-6.18$ to $-6.90\,\mathrm{kcal/mol}$ at an RMSD target of 1.0. This is likely due to Vina's direct penalty on rotatable bonds and by the easier conformational search for more rigid molecules. Full results for all runs are given in Table~\ref{tab:full-propopt-results}.


\subsection{Real-World Case Studies}
\label{subsection:case-studies}

\paragraph{Dual-Target Binder Design}

\begin{figure}[t!]
    \centering
    \includegraphics[width=\textwidth]{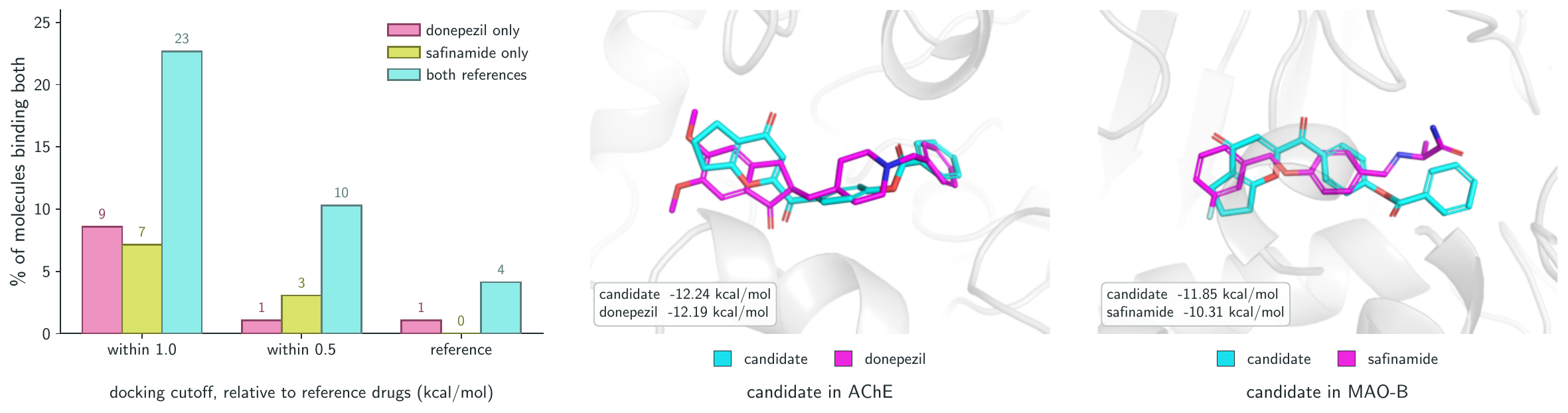}
    \caption{Dual-target binder design for AChE and MAO-B. Left: the proportion of generated molecules which bind both pockets, at cutoffs set relative to the two reference drugs' own docking scores (AChE $-12.19$, MAO-B $-10.31$ kcal/mol), when conditioning on donepezil only, safinamide only, and both references. Centre and right: the best joint candidate in its docked pose (cyan) overlaid on the crystal pose (magenta) of donepezil in AChE and of safinamide in MAO-B. Scores are from docking both the candidate and the reference drug into each pocket.}
    \label{fig:dual-binder-design}
\end{figure}

Acetylcholinesterase (AChE) inhibitors are used to treat Alzheimer's disease and monoamine oxidase B (MAO-B) inhibitors for Parkinson's; MAO-B is also implicated in Alzheimer's pathology, and dual AChE/MAO-B inhibitors are an actively pursued therapeutic strategy~\cite{medchem:alzheimers-dual-targeting} and a challenging test case for multi-objective design. Here, we apply our ensemble-conditioned guidance method to condition on the pharmacophore patterns of two reference molecules, donepezil for AChE (PDB 4EY7~\cite{medchem:ache-pdb-structure}) and safinamide for MAO-B (PDB 2V5Z~\cite{medchem:maob-pdb-structure}). Pharmacophores for each reference molecule are randomly dropped out at a rate of 20\% to encourage chemical diversity. We condition on equivariant features for donepezil and invariant for safinamide, and generate 100 molecules with $\gamma = 2.0$, with $\alpha = (0.5, 0.5)$ over the donepezil and safinamide pharmacophore conditions. Each reference is run on its own under the same setup as a single-target control, using equivariant features in both cases.

We evaluate all valid, non-fragmented molecules using AutoDock Vina with exhaustiveness 32, scoring molecules against a cutoff defined relative to the reference molecules' own docking scores (after redocking). We find that conditioning on both references produces the most dual-binding molecules (according to the docking oracle) across all thresholds (Fig.~\ref{fig:dual-binder-design}). Within $1.0\,\mathrm{kcal/mol}$ of both references, 23\% of the combined molecules bind both, against 9\% and 7\% for donepezil and safinamide alone. The gap widens as the cutoff tightens, to 10\% against 1\% and 3\% within $0.5$ kcal/mol, and 4\% against 1\% and 0\% at the reference scores themselves. The best joint candidate matches donepezil in AChE ($-12.24$ against $-12.19$ kcal/mol) and improves on safinamide in MAO-B ($-11.85$ against $-10.31$), overlaying the ring system of each drug in the corresponding pocket while remaining a novel scaffold rather than a copy of either.

\paragraph{Agonist Design}

G protein-coupled receptors (GPCRs) switch between an active and an inactive conformation, and only the active state signals. An agonist stabilises the active state while an antagonist binds the same pocket without activating the protein. This makes agonist design a problem of selectively binding the active state over the inactive, rather than maximising affinity for a single one. Here, we take the A2A adenosine receptor as a canonical example, with PDB 5G53~\cite{medchem:a2a-active-structure} as the active state bound to NECA, a known agonist, and PDB 4EIY~\cite{medchem:a2a-inactive-structure} as the inactive state bound to ZM241385, a known antagonist. Pockets are first extracted by taking residues within 6Å of any atom of the reference ligands. We align the inactive pocket onto the active and encode both with equivariant features, since the reference frames are aligned. Embedded residues with a centre-of-mass (CoM) shift of less than a given threshold between active and inactive conformations are discarded. We experiment with thresholds of 1.5Å and 2.0Å, which retain 6 residues and 2 residues, respectively. We generate 100 molecules per run with $\gamma = 2.0$ and $\alpha = (2.0, -1.0)$, and compare against a baseline conditioned on the active pocket alone. Every molecule is docked into both receptors, and we report $\Delta \mathrm{Vina}_{state} = \mathrm{Vina}_{inactive} - \mathrm{Vina}_{active}$, such that positive values refer to active state-preferring.

\begin{figure}[t!]
    \centering
    \includegraphics[width=\textwidth]{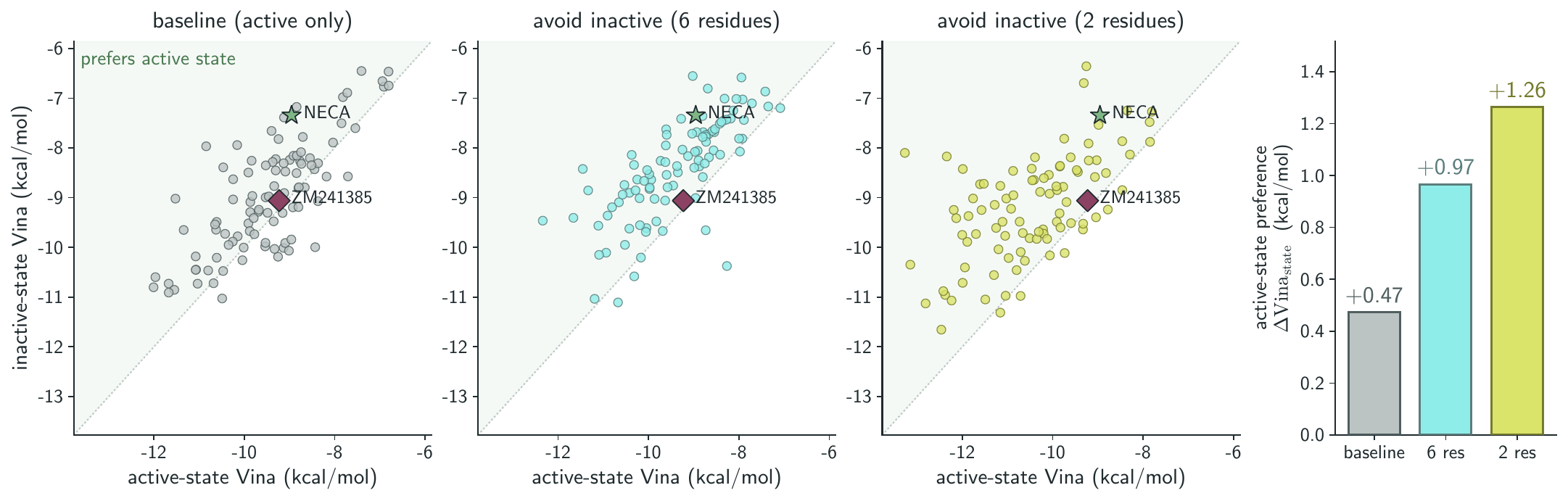}
    \caption{Active-state-biased design for the A2A adenosine receptor. Left three panels: every generated molecule placed by its docking score against the active and inactive receptor, where points above the diagonal prefer the active state, shown for the active-pocket-only baseline and for the two CoM shift thresholds used to build the inactive-pocket negative. NECA (agonist) and ZM241385 (antagonist) are docked through the same pipeline as reference points. Right: mean $\Delta \mathrm{Vina}_{state}$ for each run.}
    \label{fig:agonist-design}
\end{figure}

Adding the inactive state avoidance signal noticeably shifts generation towards the active state (Fig.~\ref{fig:agonist-design}). Conditioning generation on the active pocket alone is biased towards the active state, with a mean $\Delta \mathrm{Vina}_{state}$ of $+0.47$ kcal/mol. However, adding the inactive conformation as a negative condition roughly doubles this, giving $+0.97$ for the 6-residue condition and $+1.26$ for the 2-residue run. This compares to $+1.61$ for NECA and $+0.17$ for ZM241385 docked through the same pipeline. Since both conditions are only pocket conformations rather than binding profiles, we find that the ECFP Tanimoto to NECA stays very low throughout, with medians between $0.12$ and $0.19$, so the model is able to explore novel chemical space while still producing active state-favouring molecules.

\section{Conclusion}
\label{section:conclusion}

We proposed to approach molecular design through the modes and properties of a molecule's conformational ensemble, rather than through a single bioactive conformation, and introduced ensemble-conditioned guidance as a framework for doing so. By composing conditions at inference rather than learning them jointly, our method requires no matched multi-mode training data, and allows shapes, pharmacophores, pockets and ensemble properties to be combined in arbitrary numbers, targeted or avoided, with a single trained model. On our multi-mode benchmarks we find a clear shift towards the target shape relative to a size-matched virtual screening baseline, in both the targeting and avoidance settings, without any loss of generation quality or chemical novelty. Ensemble properties respond monotonically to the conditioning target while the conditioned binding mode is retained, showing that the two axes can be controlled concurrently even when they constrain the molecule differently. Two case studies apply the framework to practical drug design problems: for dual-target binder design, conditioning on both references substantially improves the proportion of molecules scoring well against both pockets, and for agonist design, adding the inactive receptor conformation as a negative condition more than doubles the active-state preference of generated molecules.

\paragraph{Limitations}

On the method, conditions encoded with invariant features show a weaker conditioning signal than the condition which defines the generation frame, so the assignment of the reference frame remains a meaningful choice at inference. Additionally, inference time scales linearly with the number of mode conditions since each requires its own forward pass through the generator. Property conditioning provides directional rather than calibrated control, with achieved values compressed towards the unconditioned baseline at the ends of each sweep.

On the evaluation, we rely on an approximate ensemble sampling procedure to keep our benchmarks tractable, so we cannot guarantee that generated molecules fit the desired shapes at CREST ensemble quality, and our mean PSA and mean pairwise RMSD values only approximate the CREST observables the model was trained on. CREST itself produces conformer ensembles rather than samples from a thermalized distribution, so we characterise our conditions and properties in terms of the conformational ensemble rather than making distributional claims about it. Finally, our case studies use docking scores as an oracle for binding, which is a coarse proxy and cannot distinguish designs which would succeed experimentally from those which merely score well.

\paragraph{Future Work}

A natural extension of our framework is to chameleonicity, the ability of a molecule to adopt distinct conformational states in different environments, which has been suggested as a route to designing orally bioavailable PROTACs, macrocycles and peptides~\cite{heuristics:chameleonicity-review,heuristics:psa-bro5}. Chameleonic constraints can be expressed directly as a combination of mode and property conditions, making our framework a very suitable choice, although this would require extending the maximum size of molecules beyond 48 heavy atoms.

On the methodological side, we are interested in better ways of combining vector fields, either allowing equivariant fields to be composed directly or pushing a stronger signal into the invariant features. Investigating other composition methods~\cite{multimode:dombi,multimode:superdiff} and sequential Monte Carlo approaches targeting the desired multi-modal distribution~\cite{multimode:fkc,multimode:fm-fkc} are natural candidates. Scaling up protein-ligand data with synthetic datasets would allow a wider range of pocket conditions to be learned. Validating the framework on a larger number of simultaneous reference frames would also test the generality of the composition beyond the two-mode setting studied here.

\newpage
\section*{Acknowledgements}

We thank Selma Moqvist and the rest of the Olsson group and Molecular AI department for helpful input on the work. We also thank Kento Abeywardane and Kenji Walker for valuable discussions.

This work was partially supported by the Wallenberg AI, Autonomous Systems and Software Program (WASP) funded by the Knut and Alice Wallenberg Foundation. Computational resources were provided by the National Academic Infrastructure for Supercomputing in Sweden (NAISS), funded by the Swedish Research Council. Computational resources were also provided on the Berzelius system funded by the Knut and Alice Wallenberg foundation and operated by NAISS.


\bibliographystyle{unsrtnat}
\bibliography{references}

@article{arch:adaln,
  title={Understanding and improving layer normalization},
  author={Xu, Jingjing and Sun, Xu and Zhang, Zhiyuan and Zhao, Guangxiang and Lin, Junyang},
  journal={Advances in neural information processing systems},
  volume={32},
  year={2019}
}

@InProceedings{arch:dit,
    author    = {Peebles, William and Xie, Saining},
    title     = {Scalable Diffusion Models with Transformers},
    booktitle = {Proceedings of the IEEE/CVF International Conference on Computer Vision (ICCV)},
    month     = {October},
    year      = {2023},
    pages     = {4195-4205}
}

@inproceedings{fm:gaussian-fm,
title={Flow Matching for Generative Modeling},
author={Yaron Lipman and Ricky T. Q. Chen and Heli Ben-Hamu and Maximilian Nickel and Matthew Le},
booktitle={The Eleventh International Conference on Learning Representations},
year={2023},
url={https://openreview.net/forum?id=PqvMRDCJT9t}
}

@inproceedings{fm:sto-interps-iclr,
title={Building Normalizing Flows with Stochastic Interpolants},
author={Michael Samuel Albergo and Eric Vanden-Eijnden},
booktitle={The Eleventh International Conference on Learning Representations},
year={2023},
url={https://openreview.net/forum?id=li7qeBbCR1t}
}

@inproceedings{fm:rectified-flows,
title={Flow Straight and Fast: Learning to Generate and Transfer Data with Rectified Flow},
author={Xingchao Liu and Chengyue Gong and qiang liu},
booktitle={The Eleventh International Conference on Learning Representations},
year={2023},
url={https://openreview.net/forum?id=XVjTT1nw5z}
}

@article{fm:ot-cfm,
    title={Improving and generalizing flow-based generative models with minibatch optimal transport},
    author={Alexander Tong and Kilian FATRAS and Nikolay Malkin and Guillaume Huguet and Yanlei Zhang and Jarrid Rector-Brooks and Guy Wolf and Yoshua Bengio},
    journal={Transactions on Machine Learning Research},
    issn={2835-8856},
    year={2024},
    url={https://openreview.net/forum?id=CD9Snc73AW},
    note={Expert Certification}
}

@InProceedings{fm:multiflow,
  title={Generative Flows on Discrete State-Spaces: Enabling Multimodal Flows with Applications to Protein Co-Design},
  author={Campbell, Andrew and Yim, Jason and Barzilay, Regina and Rainforth, Tom and Jaakkola, Tommi},
  booktitle={Proceedings of the 41st International Conference on Machine Learning},
  year={2024},
  publisher={PMLR},
  url={https://proceedings.mlr.press/v235/campbell24a.html}
}

@inproceedings{guide:discrete-fm,
title={Unlocking Guidance for Discrete State-Space Diffusion and Flow Models},
author={Hunter Nisonoff and Junhao Xiong and Stephan Allenspach and Jennifer Listgarten},
booktitle={The Thirteenth International Conference on Learning Representations},
year={2025},
url={https://openreview.net/forum?id=XsgHl54yO7}
}

@misc{guide:guided-flows,
  title={Guided Flows for Generative Modeling and Decision Making}, 
  author={Qinqing Zheng and Matt Le and Neta Shaul and Yaron Lipman and Aditya Grover and Ricky T. Q. Chen},
  year={2023},
  eprint={2311.13443},
  archivePrefix={arXiv},
  url={https://arxiv.org/abs/2311.13443},
}

@inproceedings{guide:classifier,
title={Diffusion Models Beat {GAN}s on Image Synthesis},
author={Prafulla Dhariwal and Alexander Quinn Nichol},
booktitle={Advances in Neural Information Processing Systems},
year={2021},
url={https://openreview.net/forum?id=AAWuCvzaVt}
}

@misc{guide:cfg,
  title={Classifier-Free Diffusion Guidance}, 
  author={Jonathan Ho and Tim Salimans},
  year={2022},
  eprint={2207.12598},
  archivePrefix={arXiv},
  url={https://arxiv.org/abs/2207.12598}
}

@inproceedings{multimode:dombi,
title={Composition of Pretrained Diffusion Models: A Logic-Based Calculus},
author={Peter Blohm and Vikas K Garg},
booktitle={The Fourteenth International Conference on Learning Representations},
year={2026},
url={https://openreview.net/forum?id=ADLiUSC7Qm}
}

@inproceedings{multimode:superdiff,
  title={The superposition of diffusion models using the it{\^o} density estimator},
  author={Skreta, Marta and Atanackovic, Lazar and Bose, Joey and Tong, Alexander and Neklyudov, Kirill},
  booktitle={International Conference on Learning Representations},
  volume={2025},
  pages={67004--67057},
  year={2025}
}

@inproceedings{multimode:fkc,
  title={Feynman-Kac Correctors in Diffusion: Annealing, Guidance, and Product of Experts},
  author={Skreta, Marta and Akhound-Sadegh, Tara and Ohanesian, Viktor and Bondesan, Roberto and Aspuru-Guzik, Alan and Doucet, Arnaud and Brekelmans, Rob and Tong, Alexander and Neklyudov, Kirill},
  booktitle={International Conference on Machine Learning},
  pages={55906--55949},
  year={2025},
  organization={PMLR}
}

@article{multimode:fm-fkc,
  title={Feynman-Kac-Flow: Inference Steering of Conditional Flow Matching to an Energy-Tilted Posterior},
  author={Mark, Konstantin and Galustian, Leonard and Kovar, Maximilian P-P and Heid, Esther},
  journal={arXiv preprint arXiv:2509.01543},
  year={2025}
}

@article{multimode:comp-gen-energy-models,
  title={Compositional visual generation with energy based models},
  author={Du, Yilun and Li, Shuang and Mordatch, Igor},
  journal={Advances in Neural Information Processing Systems},
  volume={33},
  pages={6637--6647},
  year={2020}
}

@inproceedings{multimode:comp-diff,
  title={Compositional visual generation with composable diffusion models},
  author={Liu, Nan and Li, Shuang and Du, Yilun and Torralba, Antonio and Tenenbaum, Joshua B},
  booktitle={European conference on computer vision},
  pages={423--439},
  year={2022},
  organization={Springer}
}

@inproceedings{multimode:comp-gen-mcmc,
  title={Reduce, reuse, recycle: Compositional generation with energy-based diffusion models and mcmc},
  author={Du, Yilun and Durkan, Conor and Strudel, Robin and Tenenbaum, Joshua B and Dieleman, Sander and Fergus, Rob and Sohl-Dickstein, Jascha and Doucet, Arnaud and Grathwohl, Will Sussman},
  booktitle={International conference on machine learning},
  pages={8489--8510},
  year={2023},
  organization={PMLR}
}

@inproceedings{multimode:log-diff,
title={Logical Guidance for the Exact Composition of Diffusion Models},
author={Francesco Alesiani and Jonathan H Warrell and Tanja Bien and Henrik Christiansen and Matheus Ferraz and Mathias Niepert},
booktitle={Forty-third International Conference on Machine Learning},
year={2026},
url={https://openreview.net/forum?id=OAM1jJsMGp}
}

@article{multimode:p-of-e,
  title={Training products of experts by minimizing contrastive divergence},
  author={Hinton, Geoffrey E},
  journal={Neural computation},
  volume={14},
  number={8},
  pages={1771--1800},
  year={2002},
  publisher={MIT Press}
}

@inproceedings{3dgen:eqgat-diff,
  title={Navigating the design space of equivariant diffusion-based generative models for de novo 3d molecule generation},
  author={Le, Tuan and Cremer, Julian and Noe, Frank and Clevert, Djork-Arn{\'e} and Sch{\"u}tt, Kristof T},
  booktitle={International Conference on Learning Representations},
  volume={2024},
  pages={26244--26266},
  year={2024}
}

@article{3dgen:flowr,
  title={Flowr: Flow matching for structure-aware de novo, interaction-and fragment-based ligand generation},
  author={Cremer, Julian and Irwin, Ross and Tibo, Alessandro and Janet, Jon Paul and Olsson, Simon and Clevert, Djork-Arn{\'e}},
  journal={arXiv preprint arXiv:2504.10564},
  year={2025}
}

@inproceedings{3dgen:semlaflow,
  title={SemlaFlow--Efficient 3D Molecular Generation with Latent Attention and Equivariant Flow Matching},
  author={Irwin, Ross and Tibo, Alessandro and Janet, Jon Paul and Olsson, Simon},
  booktitle={The 28th International Conference on Artificial Intelligence and Statistics},
  year={2025}
}

@article{3dgen:equifm,
  title={Equivariant flow matching with hybrid probability transport for 3d molecule generation},
  author={Song, Yuxuan and Gong, Jingjing and Xu, Minkai and Cao, Ziyao and Lan, Yanyan and Ermon, Stefano and Zhou, Hao and Ma, Wei-Ying},
  journal={Advances in Neural Information Processing Systems},
  volume={36},
  pages={549--568},
  year={2023}
}

@article{3dgen:flowmol3,
  title={FlowMol3: flow matching for 3D de novo small-molecule generation},
  author={Dunn, Ian and Koes, David R},
  journal={Digital Discovery},
  year={2026},
  publisher={Royal Society of Chemistry}
}

@inproceedings{3dgen:adit,
title={All-atom Diffusion Transformers: Unified generative modelling of molecules and materials},
author={Chaitanya K. Joshi and Xiang Fu and Yi-Lun Liao and Vahe Gharakhanyan and Benjamin Kurt Miller and Anuroop Sriram and Zachary Ward Ulissi},
booktitle={Forty-second International Conference on Machine Learning},
year={2025},
url={https://openreview.net/forum?id=89QPmZjIhv}
}

@article{3dgen:tabasco,
title={{TABASCO}: A Fast, Simplified Model for Molecular Generation with Improved Physical Quality},
author={Carlos Vonessen and Charles Harris and Miruna Cretu and Pietro Lio},
journal={Transactions on Machine Learning Research},
issn={2835-8856},
year={2026},
url={https://openreview.net/forum?id=Kg6CSrbXl4}
}

@article{3dgen:morph,
  title={Generative Molecular Morphing for Flexible-Size Design via Unbalanced Optimal Transport},
  author={Franke, Malte and Schmid, Stefan P and Ivkovic, Zarko and Jorner, Kjell and Krause, Andreas},
  journal={arXiv preprint arXiv:2606.07239},
  year={2026}
}

@inproceedings{3dgen:flexiflow,
title={FlexiFlow: decomposable flow matching for generation of flexible molecular ensemble},
author={Riccardo Tedoldi and Ola Engkvist and Patrick Bryant and Hossein Azizpour and Jon Paul Janet and Alessandro Tibo},
booktitle={Forty-third International Conference on Machine Learning},
year={2026},
url={https://openreview.net/forum?id=iL4Uo9HeXc}
}

@article{pocketcond:3d-sbdd,
  title={A 3D generative model for structure-based drug design},
  author={Luo, Shitong and Guan, Jiaqi and Ma, Jianzhu and Peng, Jian},
  journal={Advances in neural information processing systems},
  volume={34},
  pages={6229--6239},
  year={2021}
}

@inproceedings{pocketcond:pocket2mol,
  title={Pocket2mol: Efficient molecular sampling based on 3d protein pockets},
  author={Peng, Xingang and Luo, Shitong and Guan, Jiaqi and Xie, Qi and Peng, Jian and Ma, Jianzhu},
  booktitle={International conference on machine learning},
  pages={17644--17655},
  year={2022},
  organization={PMLR}
}

@inproceedings{pocketcond:target-diff,
title={3D Equivariant Diffusion for Target-Aware Molecule Generation and Affinity Prediction},
author={Jiaqi Guan and Wesley Wei Qian and Xingang Peng and Yufeng Su and Jian Peng and Jianzhu Ma},
booktitle={The Eleventh International Conference on Learning Representations },
year={2023},
url={https://openreview.net/forum?id=kJqXEPXMsE0}
}

@article{pocketcond:diff-sbdd,
  title={Structure-based drug design with equivariant diffusion models},
  author={Schneuing, Arne and Harris, Charles and Du, Yuanqi and Didi, Kieran and Jamasb, Arian and Igashov, Ilia and Du, Weitao and Gomes, Carla and Blundell, Tom L and Lio, Pietro and others},
  journal={Nature Computational Science},
  volume={4},
  number={12},
  pages={899--909},
  year={2024},
  publisher={Nature Publishing Group US New York}
}

@article{pocketcond:pilot,
  title={PILOT: equivariant diffusion for pocket-conditioned de novo ligand generation with multi-objective guidance via importance sampling},
  author={Cremer, Julian and Le, Tuan and No{\'e}, Frank and Clevert, Djork-Arn{\'e} and Sch{\"u}tt, Kristof T},
  journal={Chemical Science},
  volume={15},
  number={36},
  pages={14954--14967},
  year={2024},
  publisher={The Royal Society of Chemistry}
}

@article{pocketcond:flowr-root,
  title={FLOWR. ROOT--A flow matching-based foundation model for joint multi-purpose structure-aware 3D ligand generation and affinity prediction},
  author={Cremer, Julian and Le, Tuan and Ghahremanpour, Mohammad M and S{\l}ugocka, Emilia and Menezes, Filipe and Clevert, Djork-Arn{\'e}},
  journal={Nature Communications},
  volume={17},
  number={1},
  pages={5883},
  year={2026},
  publisher={Nature Publishing Group UK London}
}

@inproceedings{pocketcond:drug-flow,
title={Multi-domain Distribution Learning for De Novo Drug Design},
author={Arne Schneuing and Ilia Igashov and Adrian W. Dobbelstein and Thomas Castiglione and Michael M. Bronstein and Bruno Correia},
booktitle={The Thirteenth International Conference on Learning Representations},
year={2025},
url={https://openreview.net/forum?id=g3VCIM94ke}
}

@article{pocketcond:treinvent,
  title={Assessing the factors influencing the quality of pocket-conditioned 3D generative models},
  author={Wang, Kunyu and Lai, Helen and Irwin, Ross and Janet, Jon Paul and Tibo, Alessandro},
  journal={Journal of Cheminformatics},
  volume={18},
  number={1},
  pages={82},
  year={2026},
  publisher={Springer}
}

@article{benchmarking:posecheck,
  title={Benchmarking generated poses: How rational is structure-based drug design with generative models?},
  author={Harris, Charles and Didi, Kieran and Jamasb, Arian R and Joshi, Chaitanya K and Mathis, Simon V and Lio, Pietro and Blundell, Tom},
  journal={arXiv preprint arXiv:2308.07413},
  year={2023}
}

@article{benchmarking:genbench3d,
  title={Benchmarking structure-based three-dimensional molecular generative models using GenBench3D: ligand conformation quality matters},
  author={Baillif, Benoit and Cole, Jason and McCabe, Patrick and Bender, Andreas},
  journal={arXiv preprint arXiv:2407.04424},
  year={2024}
}

@article{biogen:specific-protein-shape-design,
  title={Automated design of specificity in molecular recognition},
  author={Havranek, James J and Harbury, Pehr B},
  journal={nature structural biology},
  volume={10},
  number={1},
  pages={45--52},
  year={2003},
  publisher={Nature Publishing Group US New York}
}

@article{biogen:switchable-protein-design,
  title={Computational design of a single amino acid sequence that can switch between two distinct protein folds},
  author={Ambroggio, Xavier I and Kuhlman, Brian},
  journal={Journal of the American Chemical Society},
  volume={128},
  number={4},
  pages={1154--1161},
  year={2006},
  publisher={ACS Publications}
}

@article{biogen:multistate-protein-design-framework,
  title={A generic program for multistate protein design},
  author={Leaver-Fay, Andrew and Jacak, Ron and Stranges, P Benjamin and Kuhlman, Brian},
  journal={PloS one},
  volume={6},
  number={7},
  pages={e20937},
  year={2011},
  publisher={Public Library of Science San Francisco, USA}
}

@article{biogen:multistate-protein-design-review,
  title={Multistate approaches in computational protein design},
  author={Davey, James A and Chica, Roberto A},
  journal={Protein Science},
  volume={21},
  number={9},
  pages={1241--1252},
  year={2012},
  publisher={Wiley Online Library}
}

@inproceedings{biogen:grnade,
  title={grnade: Geometric deep learning for 3d rna inverse design},
  author={Joshi, Chaitanya and Jamasb, Arian and Vi{\~n}as, Ramon and Harris, Charles and Mathis, Simon and Morehead, Alex and Anand, Rishabh and Li{\`o}, Pietro},
  booktitle={International Conference on Learning Representations},
  volume={2025},
  pages={10166--10191},
  year={2025}
}

@inproceedings{biogen:dynamic-mpnn,
title={Multi-state Protein Sequence Design with Dynamic{MPNN}},
author={Alex Abrudan and Sebastian Pujalte Ojeda and Chaitanya K. Joshi and Matthew Greenig and Felipe Engelberger and Alena Khmelinskaia and Jens Meiler and Michele Vendruscolo and Tuomas Knowles},
booktitle={The Fourteenth International Conference on Learning Representations},
year={2026},
url={https://openreview.net/forum?id=4ptHfbHG3D}
}

@article {biogen:prodit,
	author = {Jing, Bowen and Sappington, Anna and Bafna, Mihir and Shah, Ravi and Tang, Adrina and Krishna, Rohith and Klivans, Adam and Diaz, Daniel J. and Berger, Bonnie},
	title = {Generating functional and multistate proteins with a multimodal diffusion transformer},
	year = {2025},
	doi = {10.1101/2025.09.03.672144},
	publisher = {Cold Spring Harbor Laboratory},
	URL = {https://www.biorxiv.org/content/early/2025/09/04/2025.09.03.672144},
	journal = {bioRxiv}
}

@article{biogen:proteingenerator,
  title={Multistate and functional protein design using RoseTTAFold sequence space diffusion},
  author={Lisanza, Sidney Lyayuga and Gershon, Jacob Merle and Tipps, Samuel WK and Sims, Jeremiah Nelson and Arnoldt, Lucas and Hendel, Samuel J and Simma, Miriam K and Liu, Ge and Yase, Muna and Wu, Hongwei and others},
  journal={Nature biotechnology},
  volume={43},
  number={8},
  pages={1288--1298},
  year={2025},
  publisher={Nature Publishing Group US New York}
}

@inproceedings{biogen:switch-craft,
title={SwitchCraft: A Programmatic Framework for Designing State-Switching Proteins},
author={Bowen Jing and Mihir Bafna and Anisha Parsan and Heyuan Michael Ni and David Kwabi-Addo and Bryan D. Bryson and Adam Klivans and Bonnie Berger},
booktitle={Forty-third International Conference on Machine Learning},
year={2026},
url={https://openreview.net/forum?id=YtqaHnqv8c}
}

@article {biogen:caliby,
	author = {Shuai, Richard W. and Lu, Tianyu and Bhatti, Subhang and Kouba, Petr and Huang, Po-Ssu},
	title = {Ensemble-conditioned protein sequence design with Caliby},
	year = {2025},
	doi = {10.1101/2025.09.30.679633},
	publisher = {Cold Spring Harbor Laboratory},
	URL = {https://www.biorxiv.org/content/early/2025/10/05/2025.09.30.679633},
	journal = {bioRxiv}
}

@article{biogen:allogen,
  title={AlloGen: Conformation-Selective Binder Generation with Differential State Scoring},
  author={Cao, Hanqun and Quinn, Zachary and Pal, Aastha and Kimura, Sumi and Zhang, Jingjie and Heng, Pheng Ann and Chatterjee, Pranam},
  journal={arXiv preprint arXiv:2606.05474},
  year={2026}
}

@inproceedings{confgen:mcf,
title={Swallowing the Bitter Pill: Simplified Scalable Conformer Generation},
author={Yuyang Wang and Ahmed A. A. Elhag and Navdeep Jaitly and Joshua M. Susskind and Miguel {\'A}ngel Bautista},
booktitle={Forty-first International Conference on Machine Learning},
year={2024},
url={https://openreview.net/forum?id=I44Em5D5xy}
}

@article{confgen:equi-ot,
  title={Equivariant flow matching},
  author={Klein, Leon and Kr{\"a}mer, Andreas and No{\'e}, Frank},
  journal={Advances in Neural Information Processing Systems},
  volume={36},
  pages={59886--59910},
  year={2023}
}

@inproceedings{mlip:escaip,
title={The Importance of Being Scalable: Improving the Speed and Accuracy of Neural Network Interatomic Potentials Across Chemical Domains},
author={Eric Qu and Aditi S. Krishnapriyan},
booktitle={The Thirty-eighth Annual Conference on Neural Information Processing Systems},
year={2024},
url={https://openreview.net/forum?id=Y4mBaZu4vy}
}

@article{multipocket:fusediff,
  title={FuseDiff: Symmetry-Preserving Joint Diffusion for Dual-Target Structure-Based Drug Design},
  author={Wu, Jianliang and Qiao, Anjie and Wang, Zhen and Wei, Zhewei and Chen, Sheng},
  journal={arXiv preprint arXiv:2603.05567},
  year={2026}
}

@inproceedings{multipocket:combimots,
  title={CombiMOTS: Combinatorial Multi-Objective Tree Search for Dual-Target Molecule Generation},
  author={Southiratn, Thibaud and Koo, Bonil and Lu, Yijingxiu and Kim, Sun},
  booktitle={International Conference on Machine Learning},
  pages={56650--56691},
  year={2025},
  organization={PMLR}
}

@article{multipocket:evosynth,
  title={Enabling multi-target drug discovery through latent evolutionary optimization and synthesis-aware prioritization (EVOSYNTH)},
  author={Nguyen, Viet Thanh Duy and Pham, Phuc and Hy, Truong-Son},
  journal={Communications Chemistry},
  year={2026},
  publisher={Nature Publishing Group UK London}
}

@article{multipocket:molsculptor,
author = {Yanheng Li  and Xiaohan Lin  and Yize Hao  and Jun Zhang  and Yundong Wu  and Yi Qin Gao},
title = {MolSculptor: an adaptive diffusion-evolution framework enabling generative drug design for multi-target affinity and selectivity},
journal = {ChemRxiv},
year = {2025},
doi = {10.26434/chemrxiv-2025-v4758-v2},
URL = {https://chemrxiv.org/doi/abs/10.26434/chemrxiv-2025-v4758-v2}
}

@article{multipocket:dual-pocket-clm,
  title={Automated design of multi-target ligands by generative deep learning},
  author={Isigkeit, Laura and H{\"o}rmann, Tim and Schallmayer, Espen and Scholz, Katharina and Lillich, Felix F and Ehrler, Johanna HM and Hufnagel, Benedikt and B{\"u}chner, Jasmin and Marschner, Julian A and Pabel, J{\"o}rg and others},
  journal={Nature Communications},
  volume={15},
  number={1},
  pages={7946},
  year={2024},
  publisher={Nature Publishing Group UK London}
}

@article{multipocket:polygon,
  title={De novo generation of multi-target compounds using deep generative chemistry},
  author={Munson, Brenton P and Chen, Michael and Bogosian, Audrey and Kreisberg, Jason F and Licon, Katherine and Abagyan, Ruben and Kuenzi, Brent M and Ideker, Trey},
  journal={Nature Communications},
  volume={15},
  number={1},
  pages={3636},
  year={2024},
  publisher={Nature Publishing Group UK London}
}

@inproceedings{multipocket:dualdiff,
title={Reprogramming Pretrained Target-Specific Diffusion Models for Dual-Target Drug Design},
author={Xiangxin Zhou and Jiaqi Guan and Yijia Zhang and Xingang Peng and Liang Wang and Jianzhu Ma},
booktitle={The Thirty-eighth Annual Conference on Neural Information Processing Systems},
year={2024},
url={https://openreview.net/forum?id=Y79L45D5ts}
}

@article{multipocket:multipocket-rl,
  title={Finding Balance: Multiobjective Optimization in Molecular Generative Modeling},
  author={Landolfi, Laura and Catalanotti, Bruno and Janet, Jon Paul},
  journal={Journal of Chemical Information and Modeling},
  year={2026},
  publisher={ACS Publications}
}

@article{datasets:geom,
  title={GEOM, energy-annotated molecular conformations for property prediction and molecular generation},
  author={Axelrod, Simon and Gomez-Bombarelli, Rafael},
  journal={Scientific data},
  volume={9},
  number={1},
  pages={185},
  year={2022},
  publisher={Nature Publishing Group UK London}
}

@article{datasets:plinder,
  title={PLINDER: The protein-ligand interactions dataset and evaluation resource},
  author={Durairaj, Janani and Adeshina, Yusuf and Cao, Zhonglin and Zhang, Xuejin and Oleinikovas, Vladas and Duignan, Thomas and McClure, Zachary and Robin, Xavier and Studer, Gabriel and Kovtun, Daniel and others},
  journal={BioRxiv},
  pages={2024--07},
  year={2024},
  publisher={Cold Spring Harbor Laboratory}
}

@article{ff:etkdg,
  title={Better informed distance geometry: using what we know to improve conformation generation},
  author={Riniker, Sereina and Landrum, Gregory A},
  journal={Journal of chemical information and modeling},
  volume={55},
  number={12},
  pages={2562--2574},
  year={2015},
  publisher={ACS Publications}
}

@article{ff:mmff,
  title={Merck molecular force field. I. Basis, form, scope, parameterization, and performance of MMFF94},
  author={Halgren, Thomas A},
  journal={Journal of computational chemistry},
  volume={17},
  number={5-6},
  pages={490--519},
  year={1996},
  publisher={Wiley Online Library}
}

@article{ff:crest,
  title={CREST—A program for the exploration of low-energy molecular chemical space},
  author={Pracht, Philipp and Grimme, Stefan and Bannwarth, Christoph and Bohle, Fabian and Ehlert, Sebastian and Feldmann, Gereon and Gorges, Johannes and M{\"u}ller, Marcel and Neudecker, Tim and Plett, Christoph and others},
  journal={The Journal of Chemical Physics},
  volume={160},
  number={11},
  year={2024},
  publisher={AIP Publishing}
}

@article{docking:vina,
  title={AutoDock Vina: improving the speed and accuracy of docking with a new scoring function, efficient optimization, and multithreading},
  author={Trott, Oleg and Olson, Arthur J},
  journal={Journal of computational chemistry},
  volume={31},
  number={2},
  pages={455--461},
  year={2010},
  publisher={Wiley Online Library}
}

@article{docking:canon-application,
  title={A geometric approach to macromolecule-ligand interactions},
  author={Kuntz, Irwin D and Blaney, Jeffrey M and Oatley, Stuart J and Langridge, Robert and Ferrin, Thomas E},
  journal={Journal of molecular biology},
  volume={161},
  number={2},
  pages={269--288},
  year={1982},
  publisher={Elsevier}
}

@article{shape:gaussian-vol-matching,
  title={A fast method of molecular shape comparison: A simple application of a Gaussian description of molecular shape},
  author={Grant, J Andrew and Gallardo, Maria A and Pickup, Barry T},
  journal={Journal of computational chemistry},
  volume={17},
  number={14},
  pages={1653--1666},
  year={1996},
  publisher={Wiley Online Library}
}

@article{shape:gaussian-vols,
  title={A Gaussian description of molecular shape},
  author={Grant, J Andrew and Pickup, BT},
  journal={The Journal of Physical Chemistry},
  volume={99},
  number={11},
  pages={3503--3510},
  year={1995},
  publisher={ACS Publications}
}

@article{shape:rocs-application,
  title={A shape-based 3-D scaffold hopping method and its application to a bacterial protein- protein interaction},
  author={Rush, Thomas S and Grant, J Andrew and Mosyak, Lidia and Nicholls, Anthony},
  journal={Journal of medicinal chemistry},
  volume={48},
  number={5},
  pages={1489--1495},
  year={2005},
  publisher={ACS Publications}
}

@inproceedings{shape:shepherd,
title={Sh{EP}h{ERD}: Diffusing shape, electrostatics, and pharmacophores for bioisosteric drug design},
author={Keir Adams and Kento Abeywardane and Jenna Fromer and Connor W. Coley},
booktitle={The Thirteenth International Conference on Learning Representations},
year={2025},
url={https://openreview.net/forum?id=KSLkFYHlYg}
}

@article{shape:diffshape,
  title={De Novo Molecular Design via Shape-Constrained Diffusion Models},
  author={Li, Bohao and Wu, Xinyu and Cao, Yu and Lin, Jie and Zhong, Jingpeng and Chen, Hua and Lu, Yongzhi and Tang, Miru and Lei, Jinping and Ran, Ting and others},
  journal={Journal of Chemical Information and Modeling},
  year={2026},
  publisher={ACS Publications}
}

@article{shape:shapemol,
  title={Shape-conditioned 3D molecule generation via equivariant diffusion models},
  author={Chen, Ziqi and Peng, Bo and Parthasarathy, Srinivasan and Ning, Xia},
  journal={arXiv preprint arXiv:2308.11890},
  year={2023}
}

@inproceedings{shape:squid,
title={Equivariant Shape-Conditioned Generation of 3D Molecules for Ligand-Based Drug Design},
author={Keir Adams and Connor W. Coley},
booktitle={The Eleventh International Conference on Learning Representations },
year={2023},
url={https://openreview.net/forum?id=4MbGnp4iPQ}
}

@article{propopt:decaf,
  title={Boltzmann-Expected Molecular Design with Decoupled Annealing Flows},
  author={Moqvist, Selma and Beckmann, Richard and Irwin, Ross and Mercado, Roc{\'\i}o and Olsson, Simon},
  journal={arXiv preprint arXiv:2607.19519},
  year={2026}
}

@inproceedings{propopt:flips,
title={Flexibility-conditioned protein structure design with flow matching},
author={Vsevolod Viliuga and Leif Seute and Nicolas Wolf and Simon Wagner and Arne Elofsson and Jan St{\"u}hmer and Frauke Gr{\"a}ter},
booktitle={Forty-second International Conference on Machine Learning},
year={2025},
url={https://openreview.net/forum?id=890gHX7ieS}
}

@article{heuristics:veber,
  title={Molecular properties that influence the oral bioavailability of drug candidates},
  author={Veber, Daniel F and Johnson, Stephen R and Cheng, Hung-Yuan and Smith, Brian R and Ward, Keith W and Kopple, Kenneth D},
  journal={Journal of medicinal chemistry},
  volume={45},
  number={12},
  pages={2615--2623},
  year={2002},
  publisher={ACS Publications}
}

@article{heuristics:flex-psa,
  title={Influence of molecular flexibility and polar surface area metrics on oral bioavailability in the rat},
  author={Lu, Jing J and Crimin, Kimberly and Goodwin, Jay T and Crivori, Patrizia and Orrenius, Christian and Xing, Li and Tandler, Peter J and Vidmar, Thomas J and Amore, Benny M and Wilson, Alan GE and others},
  journal={Journal of medicinal chemistry},
  volume={47},
  number={24},
  pages={6104--6107},
  year={2004},
  publisher={ACS Publications}
}

@article{heuristics:psa-bro5,
  title={Impact of dynamically exposed polarity on permeability and solubility of chameleonic drugs beyond the rule of 5},
  author={Rossi Sebastiano, Matteo and Doak, Bradley C and Backlund, Maria and Poongavanam, Vasanthanathan and Over, Björn and Ermondi, Giuseppe and Caron, Giulia and Matsson, Pär and Kihlberg, Jan},
  journal={Journal of Medicinal Chemistry},
  volume={61},
  number={9},
  pages={4189--4202},
  year={2018},
  publisher={ACS Publications}
}

@article{heuristics:conf-entropy-binding,
  title={Ligand configurational entropy and protein binding},
  author={Chang, Chia-en A and Chen, Wei and Gilson, Michael K},
  journal={Proceedings of the National Academy of Sciences},
  volume={104},
  number={5},
  pages={1534--1539},
  year={2007},
  publisher={National Academy of Sciences}
}

@article{heuristics:chameleonicity-review,
  title={Molecular chameleons in drug discovery},
  author={Poongavanam, Vasanthanathan and Wieske, Lianne HE and Peintner, Stefan and Erd{\'e}lyi, M{\'a}t{\'e} and Kihlberg, Jan},
  journal={Nature Reviews Chemistry},
  volume={8},
  number={1},
  pages={45--60},
  year={2024},
  publisher={Nature Publishing Group UK London}
}

@article{medchem:binding-conf-changes,
  title={Conformational changes of small molecules binding to proteins},
  author={Nicklaus, Marc C and Wang, Shaomeng and Driscoll, John S and Milne, George WA},
  journal={Bioorganic \& medicinal chemistry},
  volume={3},
  number={4},
  pages={411--428},
  year={1995},
  publisher={Elsevier}
}

@article{medchem:binding-conf-changes-2,
  title={Conformational energy penalties of protein-bound ligands},
  author={Bostr{\"o}m, Jonas and Norrby, Per-Ola and Liljefors, Tommy},
  journal={Journal of computer-aided molecular design},
  volume={12},
  number={4},
  pages={383--383},
  year={1998},
  publisher={Springer}
}

@article{medchem:binding-conf-changes-3,
  title={Conformational analysis of drug-like molecules bound to proteins: an extensive study of ligand reorganization upon binding},
  author={Perola, Emanuele and Charifson, Paul S},
  journal={Journal of medicinal chemistry},
  volume={47},
  number={10},
  pages={2499--2510},
  year={2004},
  publisher={ACS Publications}
}

@article{medchem:alzheimers-dual-targeting,
  title={Latest advances in dual inhibitors of acetylcholinesterase and monoamine oxidase B against Alzheimer’s disease},
  author={Zou, Dajiang and Liu, Renzheng and Lv, Yangjing and Guo, Jianan and Zhang, Changjun and Xie, Yuanyuan},
  journal={Journal of Enzyme Inhibition and Medicinal Chemistry},
  volume={38},
  number={1},
  pages={2270781},
  year={2023},
  publisher={Taylor \& Francis}
}

@article{medchem:ache-pdb-structure,
  title={Structures of human acetylcholinesterase in complex with pharmacologically important ligands},
  author={Cheung, Jonah and Rudolph, Michael J and Burshteyn, Fiana and Cassidy, Michael S and Gary, Ebony N and Love, James and Franklin, Matthew C and Height, Jude J},
  journal={Journal of medicinal chemistry},
  volume={55},
  number={22},
  pages={10282--10286},
  year={2012},
  publisher={ACS Publications}
}

@article{medchem:maob-pdb-structure,
  title={Structures of human monoamine oxidase B complexes with selective noncovalent inhibitors: safinamide and coumarin analogs},
  author={Binda, Claudia and Wang, Jin and Pisani, Leonardo and Caccia, Carla and Carotti, Angelo and Salvati, Patricia and Edmondson, Dale E and Mattevi, Andrea},
  journal={Journal of medicinal chemistry},
  volume={50},
  number={23},
  pages={5848--5852},
  year={2007},
  publisher={ACS Publications}
}

@article{medchem:a2a-active-structure,
  title={Structure of the adenosine A2A receptor bound to an engineered G protein},
  author={Carpenter, Byron and Nehm{\'e}, Rony and Warne, Tony and Leslie, Andrew GW and Tate, Christopher G},
  journal={Nature},
  volume={536},
  number={7614},
  pages={104--107},
  year={2016},
  publisher={Nature Publishing Group UK London}
}

@article{medchem:a2a-inactive-structure,
  title={Structural basis for allosteric regulation of GPCRs by sodium ions},
  author={Liu, Wei and Chun, Eugene and Thompson, Aaron A and Chubukov, Pavel and Xu, Fei and Katritch, Vsevolod and Han, Gye Won and Roth, Christopher B and Heitman, Laura H and IJzerman, Adriaan P and others},
  journal={Science},
  volume={337},
  number={6091},
  pages={232--236},
  year={2012},
  publisher={American Association for the Advancement of Science}
}


\newpage
\appendix


\section{Additional Related Work}
\label{section:related}

Since we consider this work to lie at the intersection of numerous approaches and ideas in molecular design, we provide a thorough discussion on related strands of work in this section.

\paragraph{Pocket-conditioned Generation}

Models which can directly generate ligands conditioned on target protein pockets have received significant attention lately. Early approaches used either autoregressive generation~\cite{pocketcond:3d-sbdd,pocketcond:pocket2mol} or diffusion generative models~\cite{pocketcond:target-diff,pocketcond:diff-sbdd,pocketcond:pilot} but often suffered from slow sampling times and chemically unrealistic samples~\cite{benchmarking:posecheck,benchmarking:genbench3d}. Later work combined flow matching with efficient equivariant architectures to significantly improve sample time and quality~\cite{3dgen:semlaflow,pocketcond:drug-flow,3dgen:flowr,pocketcond:flowr-root,pocketcond:treinvent}. Recently, FlexiFlow~\cite{3dgen:flexiflow} was introduced to allow sampling multiple bound ligand conformers within a pocket. Generating molecules of arbitrary sizes, however, remains a limitation of diffusion and flow models, although some recent work~\cite{3dgen:flowmol3,3dgen:morph} has investigated allowing flexible size generation, but these have not been extended to pocket-conditioned generation.

\paragraph{Shape Conditioning}

An orthogonal line of work to pocket conditioning has followed a similar strategy to ligand-based virtual screening, where ligands are selected based on having high Gaussian volume overlap with a reference binder~\cite{shape:gaussian-vols,shape:gaussian-vol-matching}. Instead of screening chemical space for potential binders, recent work has trained generative models to design molecules with low-energy conformers which fit the given reference shape~\cite{shape:shapemol,shape:squid,shape:diffshape}. ShEPhERD~\cite{shape:shepherd} takes this idea one step further by allowing generation conditioned on shape, pharmacophores and electrostatics. However, all of these approaches only consider generation based on a single reference (bound) shape profile.

\paragraph{Multi-target Design}

Designing molecules for multiple targets is also of significant interest. Ligand-based methods learn from compounds already associated with a target pair, either by fine-tuning chemical language models on known dual-active ligands~\cite{multipocket:dual-pocket-clm} or by reinforcement learning over an embedded chemical space~\cite{multipocket:polygon}, and so depend on prior ligand knowledge for that pair without using explicit structural information. Other approaches, such as MolSculptor~\cite{multipocket:molsculptor} and EvoSynth~\cite{multipocket:evosynth}, optimise within the latent space of a pretrained generative model and score candidates using 3D-aware surrogates. MolSculptor in particular supports both dual-target and selectivity design. CombiMOTS~\cite{multipocket:combimots} instead searches combinatorially over fragments, and, like the latent optimisation methods, does not generate directly in 3D. Structure-based methods, such as those proposed by \citet{multipocket:dualdiff}, reprogram pretrained single-target diffusion models zero-shot by aligning two pockets under a rigid transformation and composing their scores, which avoids paired training data but assumes the two binding modes are related by a single rigid motion. FuseDiff~\cite{multipocket:fusediff} drops that assumption by jointly generating two pocket-specific poses over a shared graph, but requires the same ligand resolved in both pockets, which severely limits the available training data.

\paragraph{Multi-state Biomolecular Design}

Multi-state design has a long history in computational protein design. \citet{biogen:specific-protein-shape-design} introduced explicit negative design into an automated design algorithm, selecting sequences which maximise the transfer free energy from a target conformation to a set of undesired competitor conformations, and~\citet{biogen:switchable-protein-design} optimised a single sequence against multiple target structures to produce a peptide which switches fold in response to pH or transition metals. Later work generalised this to arbitrary sets of states and scoring terms~\cite{biogen:multistate-protein-design-framework}. \citet{biogen:multistate-protein-design-review} surveys the algorithmic and scoring approaches developed in this line of work. There has also been a recent surge of interest in deep learning approaches for designing biomolecules which fit multiple states. DynamicMPNN~\cite{biogen:dynamic-mpnn} was introduced to allow the design of proteins adopting multiple conformations, although it is limited to training on paired conformations. ProDiT~\cite{biogen:prodit} proposes a method to denoise two protein structures in parallel with a single sequence, circumventing the lack of paired training data, while ProteinGenerator~\cite{biogen:proteingenerator} similarly ties together diffusion trajectories with distinct structural constraints by averaging their sequence logits. Caliby~\cite{biogen:caliby} and gRNAde~\cite{biogen:grnade}, respectively, allow protein and RNA sequence design from multiple desired conformational states. Neither supports state-avoidance, nor conditions on properties of the ensemble as a whole. SwitchCraft~\cite{biogen:switch-craft} instead treats a frozen structure prediction model as a differentiable loss, optimising a sequence against compositional constraints defined over several states, but requires a separate optimisation run for each design. Recently, AlloGen~\cite{biogen:allogen} was proposed as a method for designing conformationally-selective peptide binders, but focuses only on optimising for state-selectivity.

\paragraph{Multi-conditional Guidance}

Combining distributions multiplicatively rather than additively originates with the product of experts formalism~\cite{multimode:p-of-e}, where the joint density is the normalised product of per-expert densities and a sample is likely only if every expert assigns it high density. \citet{multimode:comp-gen-energy-models} applied this idea to compose independently trained energy functions over visual concepts, including negated ones. \citet{multimode:comp-diff} extended the construction to diffusion models by summing the score estimates of pretrained conditional models, providing conjunction and negation operators. LogDiff~\cite{multimode:log-diff} extends these to additional logical composition operators such as disjunction and exclusive-or. Since summing scores does not sample the intended intermediate marginals, later work adds corrector steps: MCMC transitions targeting the composed density at each noise level~\cite{multimode:comp-gen-mcmc}, and sequential-Monte-Carlo reweighting for annealed, geometric-averaged, and product distributions~\cite{multimode:fkc,multimode:fm-fkc}.

\paragraph{Ensemble Property Control}

Optimising properties calculated over 3D ensembles is a crucial aspect of molecular design. Heuristics based on 3D polar surface area and molecular flexibility have long served as a guide for improving bioavailability~\cite{heuristics:veber,heuristics:flex-psa}, and controlling flexibility and conformational entropy is likewise important for optimising binding affinity~\cite{heuristics:conf-entropy-binding}. Despite this, relatively little existing work addresses ensemble property optimisation in small molecule design, since most property-guided generative models tie each property to a single generated conformer. One notable recent exception is DECAF~\cite{propopt:decaf}, which produces molecules with target ensemble property values through iterative optimisation of a population of candidate molecules. Additionally, in protein design, FliPS~\cite{propopt:flips} conditions backbone generation on per-residue flexibility profiles, using a learned flexibility predictor, although conditioning is restricted to that single property. Neither DECAF nor FliPS support conditioning on specific conformational states.


\section{Extended Methods}
\label{appendix:extended-methods}

Here we provide full additional details on: the pharmacophore definitions we used for both ligand-only and protein-ligand data; further details and hyperparameters for our neural network architecture; and details on how we split our datasets into training and test subsets.

\subsection{Pharmacophore Extraction}
\label{appendix:extended-methods-pharmacophores}

Pharmacophores and protein-ligand interactions form a core part of our model's conditioning information. Since we train our model on both ligand-only (GEOM Drugs) and protein-ligand (SPINDR) data, we define separate workflows for extracting pharmacophore information but combine them in a way that preserves conditioning information. Crucially, since high-quality protein-ligand data is very limited, this allows us to significantly expand the diversity of pharmacophore patterns the model sees during training.

Pharmacophores are extracted from ligand-only data using a set of SMARTS patterns. These broadly follow the patterns used by ShEPhERD. However, to help align the pharmacophores from SMARTS with those used for protein-ligand systems, we use ShEPhERD's SMARTS rules with the following adaptations:
\begin{itemize}
    \item Remove aromatic systems from the hydrophobe group. Aromatic rings get assigned an \textit{aromatic} pharmacophore tag anyway, so this information is mostly redundant.
    \item Remove ShEPhERD's halogen group since different halogen atoms can provide very different interactions. We include bromine and iodine under our \textit{hydrophobe} group, and exclude chlorine and fluorine completely.
    \item Exclude some of ShEPhERD's hydrophobe SMARTS to align with ProLIF's hydrophobe definitions. We found that ShEPhERD's definitions tended to generate many redundant hydrophobe tags for chains of carbon atoms.
    \item Remove the zinc binder group completely since these are less relevant for drug-like molecules.
\end{itemize}

The full set of pharmacophore tags extracted from our SMARTS rules are: \textit{hydrogen bond donor}, \textit{hydrogen bond acceptor}, \textit{cation}, \textit{anion}, \textit{aromatic}, and \textit{hydrophobe}.

Although the SPINDR dataset contains pre-computed protein-ligand interactions, we re-process the systems using ProLIF for completeness. When mapping ProLIF interactions to their SMARTS counterparts we first run the ligand-only SMARTS pharmacophore extraction logic on the ligand, and then, for each ProLIF interaction, find its best matching SMARTS pharmacophore, dropping any interaction that isn't matched. Each ProLIF interaction must exactly match its corresponding pharmacophore group tag (as provided above) based on the following mapping:
\begin{itemize}
    \item ProLIF's \textit{HBDonor} and \textit{HBAcceptor} are mapped to the \textit{hydrogen bond donor} and \textit{hydrogen bond acceptor} groups, respectively.
    \item ProLIF's \textit{Cationic} and \textit{Anionic} are mapped to \textit{cation} and \textit{anion}, respectively.
    \item ProLIF's \textit{Hydrophobic} maps to \textit{hydrophobe}.
    \item ProLIF's \textit{PiStacking} and \textit{PiCation} map to \textit{aromatic}.
    \item ProLIF's \textit{CationPi} maps to \textit{cation}.
\end{itemize}

\subsection{Architecture}
\label{appendix:arch-details}

\begin{figure}[t]
    \begin{subfigure}{0.48\textwidth}
        \centering
        \includegraphics[width=\linewidth]{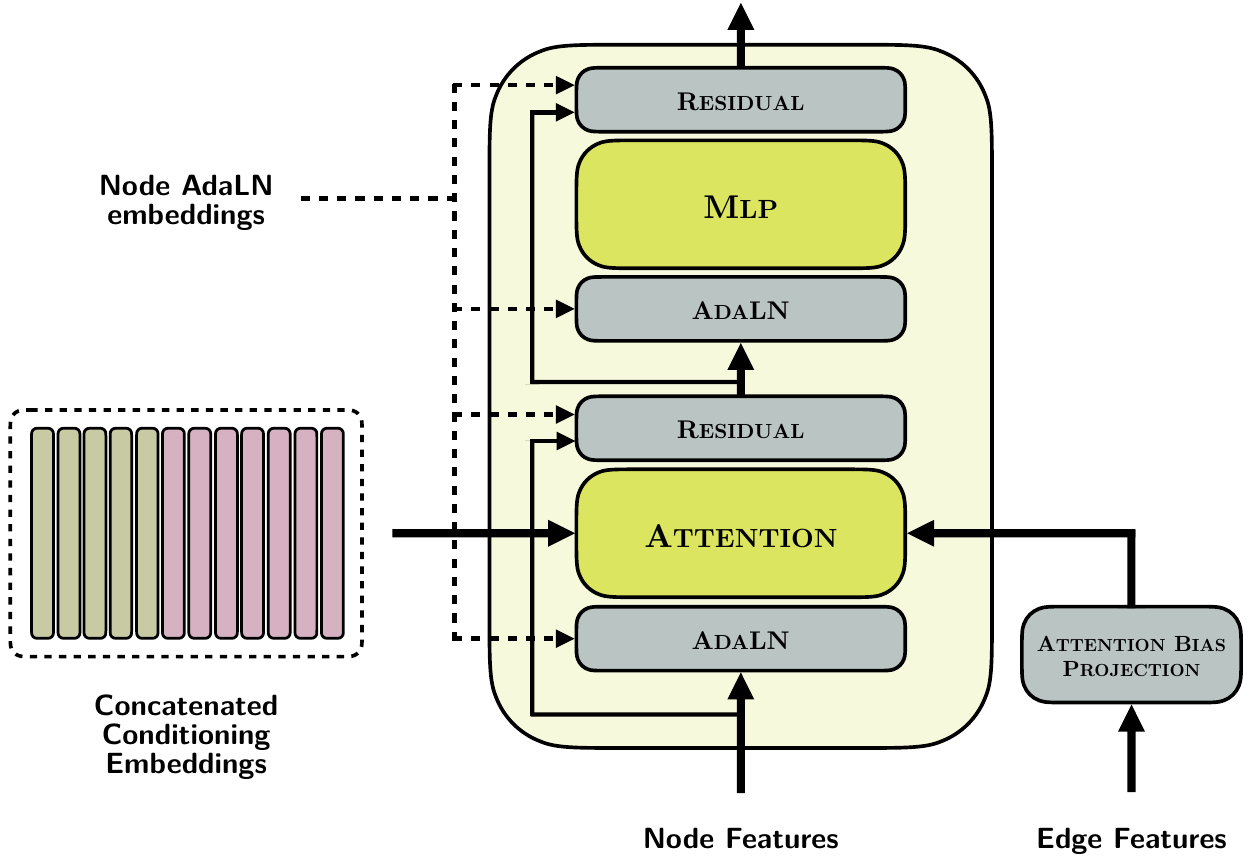} 
    \end{subfigure}
    \hfill
    \begin{subfigure}{0.44\textwidth}
        \centering
        \includegraphics[width=\linewidth]{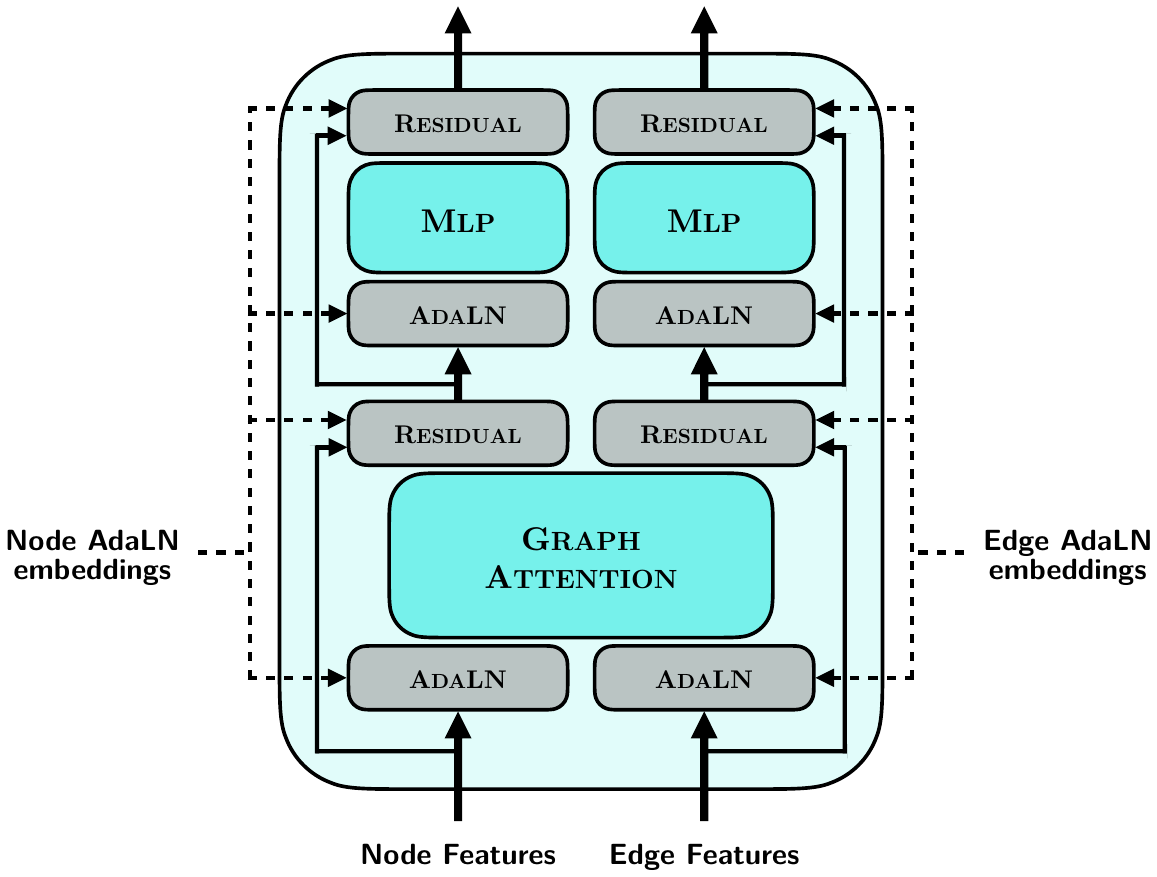}
    \end{subfigure}
    \caption{A transformer layer (left) and graph transformer layer (right), as used in the generator.}
    \label{fig:model-layers}
\end{figure}

This section provides the architectural details deferred from Section~\ref{subsection:arch}, including the encoder modules and the transformer and graph transformer layers used in the generator. Figure~\ref{fig:model-layers} illustrates both layer types.

\paragraph{Encoders}

Both the profile and pocket encoders are stacks of standard (non-equivariant) transformer layers combined with AdaLN~\cite{arch:adaln}. Encoder inputs comprise the relevant point cloud features (coordinates, types and, for the profile encoder, direction vectors). The adaptive symmetry rotated flag described in Section~\ref{subsection:arch}, along with level of shape profile noise $\sigma_{shape}$ (only for the profile encoder), are used as conditioning for AdaLN parameters.

\paragraph{Decoder Transformer Layer}

The decoder uses a similar transformer layer structure as the encoders, with three alterations. First, the attention mechanism includes conditioning embeddings (of arbitrary length), where attention queries are linear projections of the current decoder node features, while keys and values are projections of node features concatenated with the encoded profile and pocket embeddings (denoted $m$). Second, we apply an additive bias to dot-product attention scores, where biases are linear projections of incoming pairwise features. Finally, AdaLN is conditioned on the concatenated time embedding and property embedding $\eta$.

\paragraph{Graph Transformer Layer}

These layers do not see the encoded conditioning embeddings $m$, operating only on node and edge features of the partially denoised graph (Fig.~\ref{fig:model-layers}, right). The graph attention module operates similarly to SemlaFlow's latent attention~\cite{3dgen:semlaflow}, where node features are first projected into lower dimensional queries and keys. Raw attention scores between a query \(q_i\) and key \(k_j\) are computed using a gated outer product
\begin{equation}
    a_{ij} = w^\top \bigl( (q_i k_j^\top) \odot \sigma(e_{ij}) \bigr)
\end{equation}
where \(e_{ij}\) are incoming edge features and $\sigma$ is the sigmoid function. Feature accumulation then proceeds in the same way as regular dot-product attention, including the use of multiple attention heads. After graph attention, both node and edge features are updated using separate multi-layer perceptrons (MLPs).

\paragraph{Hyperparameters}

\begin{table}[t]
    \centering
    \caption{Hyperparameters for different components of the our model.}
    \label{tab:hparams}
    \begin{tabular}{lcccc}
        \toprule
        Hyperparameter & Property Encoder & Profile Encoder & Pocket Encoder & Generator \\
        \midrule
        Number of layers & ---   & 8    &  8   & $4 \times 4 = 16$  \\
        Hidden dimension & 128   & 128  & 128  & 384  \\
        Edge Dimension   & ---   & ---  & ---  & 64  \\
        Attention heads  & ---   & 8    & 8    & 16  \\
        Dropout          & 0.0   & 0.1  & 0.1  & 0.1  \\
        Parameters       & 0.08M & 2.5M & 2.5M & 53.5M  \\
        \bottomrule
    \end{tabular}
\end{table}

Table~\ref{tab:hparams} lists the hyperparameters and model sizes for different parts of our model. The total number of learnable parameters (not including EMA-updated weights) is approximately 58.6M. As outlined in Section~\ref{subsection:model-training}, we train end-to-end with the Adam optimiser (learning rate $10^{-3}$, AMSGrad, no weight decay). We use a linear warm-up of 10K steps to a constant learning rate, gradient clipping at $1.0$, and an exponential moving average of the generator weights with decay $0.999$. The EMA-averaged weights are used during evaluation. Mixed-precision (bf16) is used for both training and inference.

\subsection{Dataset Splits}
\label{appendix:data-splits}

Here we describe our strategy for producing train/val/test dataset splits. For GEOM Drugs, we opted to create a new split based on novel, unique scaffolds, to test generalisability. SPINDR splits follow those in~\citet{3dgen:flowr}.

\paragraph{GEOM Drugs}

Data splitting proceeds by first finding all molecules with a unique Murcko scaffold in the dataset (implemented using RDKit). From this set, we look for flexible, drug-like molecules by applying the following filters:
\begin{enumerate}
    \item Number of heavy atoms between 16 and 35 (inclusive).
    \item Maximum CLogP (logarithm of the octanol-water partition coefficient, estimated using RDKit) of 5.0.
    \item At least 10 conformers in the pre-computed CREST ensemble.
\end{enumerate}
We then randomly sample 1000 molecules from this filtered set as our test set. After removing the test molecules from the full set, we randomly sample 10K molecules as an additional held-out validation set, with the remaining molecules used for training.

\paragraph{SPINDR}

The protein-ligand data uses the same train and test splits used by SPINDR~\cite{3dgen:flowr}, which follow the splits originally proposed for the full PLINDER dataset~\cite{datasets:plinder}. These splits enforce strict conditions on pocket and ligand similarity between train and test data, including dropping some systems from train and test data to ensure a similarity gap. Full details can be found in~\citet{datasets:plinder}. We remove any system where the reference ligand has a QED score < 0.3 from both training and test sets.


\section{Benchmark Construction}

This section discusses the ensemble sampling algorithm used for our benchmark evaluations, as well as specific details on the setup of the mode targeting, mode avoidance and ensemble property optimisation benchmarks.

\subsection{Ensemble Sampling}
\label{appendix:ensemble-sampling}

\begin{figure}[t]
    \centering
    \includegraphics[width=\linewidth]{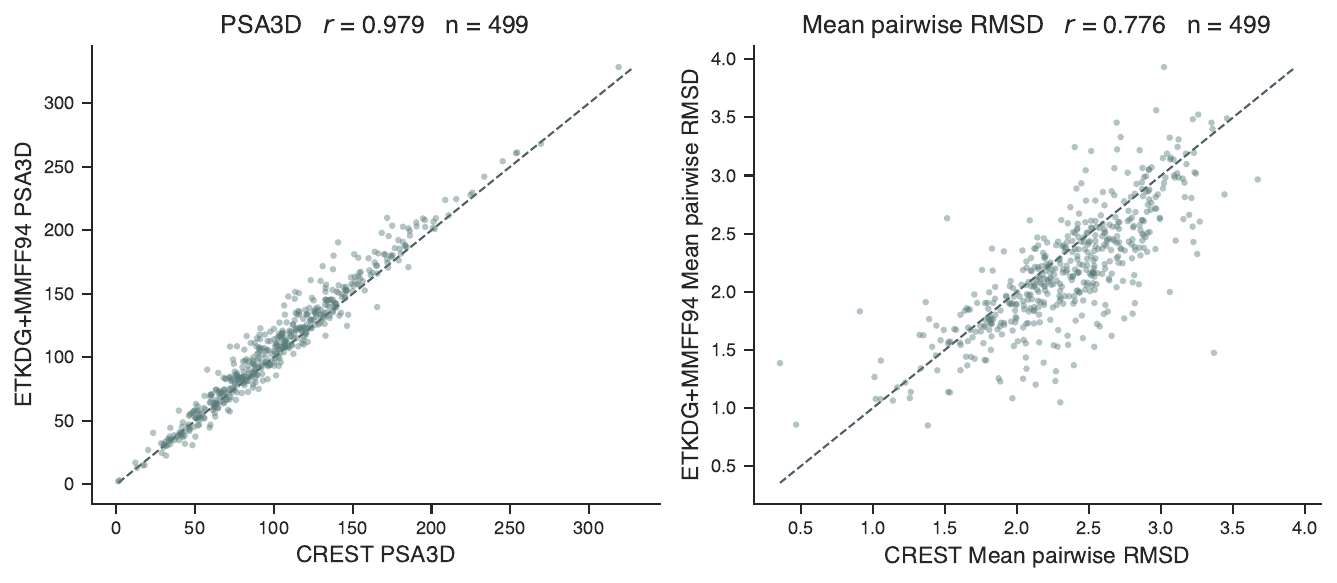}
    \caption{Scatter plots comparing mean PSA (left) and mean pairwise RMSD (right) for ensembles computed using CREST (x-axis) and our fast ensemble approximation approach (y-axis).}
    \label{fig:approx-ensemble-props}
\end{figure}

\paragraph{Ensemble Approximation}

Since we wish to allow efficient evaluation of molecules generated by our model, we set up a fast ensemble sampling approach, bypassing the use of CREST~\cite{ff:crest}, which typically takes hours per small molecule. For each molecule we first sample 128 conformers using ETKDG~\cite{ff:etkdg} (implemented in RDKit), and then run minimisation with the MMFF94 forcefield~\cite{ff:mmff} for up to 1000 steps. We then deduplicate the conformer set by iterating the list of samples and remove any with an RMSD within 0.5Å to any previously selected conformer. Finally, we calculate the MMFF energies for each remaining sample and remove any conformer where the energy difference to the minimum is more than 6.0 kcal/mol. We found that increasing the number of sampled conformers to 1024 only marginally changed the estimated values of the observables, so we use 128 conformers to keep the benchmark evaluation tractable.

\paragraph{Approximate Ensemble Properties}

In practice, for ensemble property estimation, we find this method of sampling ensembles serves as a reasonable approximation to the underlying CREST conformers. We sampled 512 molecules at random from the validation split, sampled their ETKDG+MMFF ensembles using the above approach, and calculated mean PSA and mean pairwise RMSD property values over the CREST and approximated molecular ensembles. Ensembles and properties were successfully computed for 499 of the molecules. Scatter plots comparing the properties of the two ensemble sampling approaches for these molecules are shown in Fig.~\ref{fig:approx-ensemble-props}. The Mean PSA for each molecule is almost identical between the CREST and approximated ensembles, while the mean pairwise RMSD still shows a strong correlation (Pearson R $= 0.78$).

\subsection{Multi-Mode Conditioning Benchmark}
\label{appendix:multi-cond-benchmark}

Our mode conditioning benchmark aims to test the model's ability to condition on multiple modes of the desired conformational ensemble, while retaining tractable evaluation. For simplicity we restrict the benchmark to conditioning on two modes, and use shape-only conditioning throughout. The setup begins by resampling conformer ensembles for each molecule in our test set using the procedure in Appendix~\ref{appendix:ensemble-sampling} to ensure the conditioning shapes are accessible for evaluation using the same procedure. The compact and extended conformers are then taken at the 20th and 80th percentiles (in order to avoid outliers) of radius of gyration. Pairings are cross-molecule, taking the compact conformer of one test molecule as the first mode and the extended conformer of another as the second, so that no single molecule trivially satisfies both. A pairing is only kept if we can demonstrate that it is achievable. To do this we build a pool of training molecules, sampled evenly across heavy-atom counts and given the same ensemble treatment, and score each against both target shapes using its best conformer. For mode targeting, we require at least one training molecule reaching a shape Tanimoto of 0.8 against both targets. For mode avoidance we require a shape constraint in both directions; one molecule must match the compact target at 0.8 while staying below 0.6 against the extended target, and another must do the reverse, so that the pairing remains achievable whichever mode is negated. We further require any such molecule to have an ECFP Tanimoto of at most 0.5 to both source molecules, so that a pairing is kept only when it can be satisfied by chemistry which is distinct from either source and the benchmark cannot be solved by reproducing a source molecule. These training molecules are used only to establish feasibility during construction and are never provided to the model. The size-matched baseline for each target is computed from the same pool, averaging over all molecules within one heavy atom of the reference with no chemical filter applied.

\subsection{Evaluation Metrics}
\label{appendix:eval-metrics}

This section gives full definitions of the metrics used in Section~\ref{section:results} and Appendix~\ref{appendix:additional-results}. Unless stated otherwise all reported values are means over the benchmark systems, and any metric involving a reference is computed per system against that system's reference molecule.

\begin{itemize}
    \item \textbf{Validity} The proportion of generated molecules which can be sanitised by RDKit and consist only of a single fragment. We use \textit{valid} to refer to connected validity throughout, where molecules that contain disconnected fragments are counted as failures.

    \item \textbf{Uniqueness} The proportion of distinct canonical SMILES among all generated molecules which can be converted to SMILES.

    \item \textbf{ECFP Tanimoto} Tanimoto similarity between the ECFP4 fingerprints (Morgan, radius 2, 2048 bits) of the generated and reference molecules, computed after removing hydrogens. We use this as a novelty measure, where lower values indicate that the model is not simply reproducing the reference.

    \item \textbf{Shape Tanimoto} Gaussian shape overlap between a generated and a reference conformer, computed using RDKit's \texttt{rdShapeAlign}. For the multi-mode benchmarks we sample an MMFF ensemble for each generated molecule (Appendix~\ref{appendix:ensemble-sampling}), align every conformer to the reference and keep the best scoring one.

    \item \textbf{Shape matching gain ($\Delta$)} The shape Tanimoto of a generated molecule against a target, minus a size-matched baseline for that target. The baseline is the mean shape Tanimoto over a random sample of training molecules within one heavy atom of the reference, scored in exactly the same way. $\Delta$ therefore measures the gain over virtual screening molecules of the same size.

    \item \textbf{Interaction recovery} The proportion of conditioned reference interactions for which the generated molecule places a pharmacophore of the same type within 2Å. Only the pharmacophores given to the model as conditioning are counted, so the metric measures conditioning fidelity rather than general interaction quality. It is computed on the locally relaxed conformer, as described in Section~\ref{subsection:prop-opt-expts}.

    \item \textbf{PSA and mean pairwise RMSD} The two ensemble properties defined in Section~\ref{subsection:ensemble-conditions}, computed over an MMFF ensemble sampled for each generated molecule using the procedure in Appendix~\ref{appendix:ensemble-sampling}.

    \item \textbf{Vina score and Vina min.} AutoDock Vina scores for the generated pose in its pocket, in kcal/mol, using a box centred on the reference ligand with 8Å of padding. \textit{Vina} evaluates the pose exactly as generated, while \textit{Vina min.} evaluates it after local minimisation under the Vina scoring function. Neither uses redocking, so the generated binding mode is preserved in both cases.
\end{itemize}


\section{Additional Results}
\label{appendix:additional-results}

\subsection{Size Distribution Learning}
\label{appendix:size-learning}

\begin{figure}[t]
    \centering
    \includegraphics[width=\textwidth]{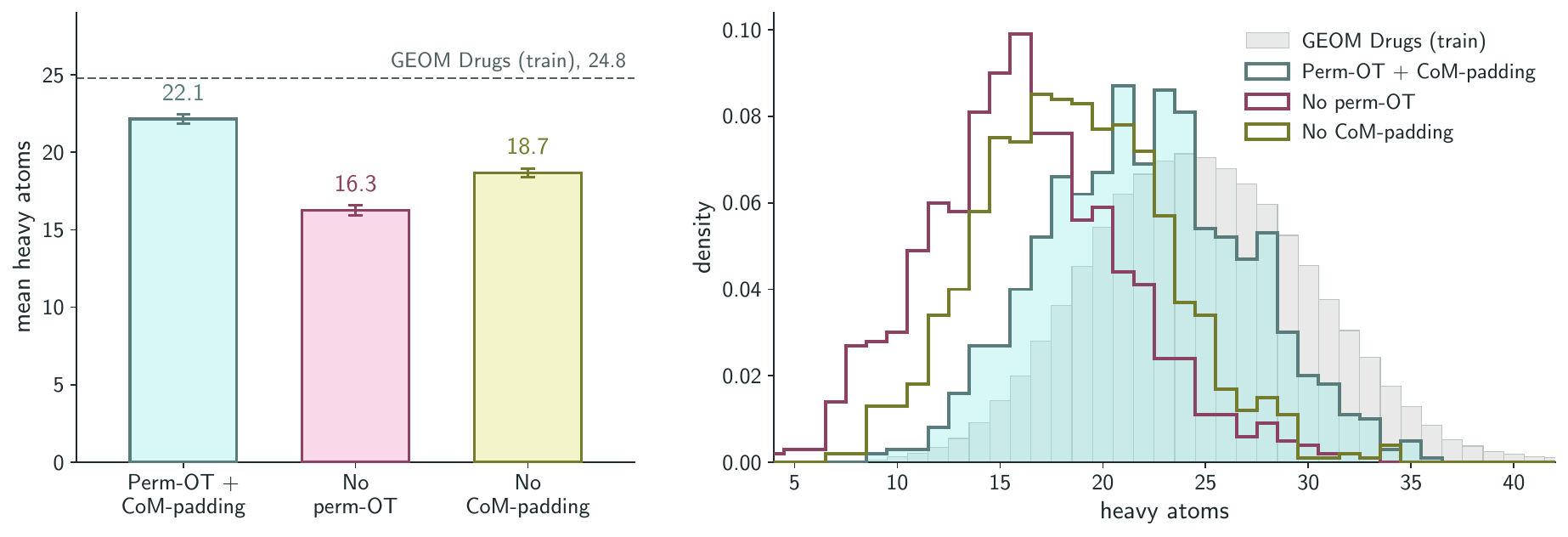}
    \caption{Unconditional size distributions for the three training setups, all trained on GEOM Drugs without the pocket encoder. \textbf{Left}: mean generated size, with the GEOM Drugs training mean marked. \textbf{Right}: full generated size distributions against the training distribution. Sizes are taken from the model's pad token predictions. Error bars are 95\% percentile bootstrap intervals over 2000 resamples of the 1000 molecules generated per model.}
    \label{fig:size-ablation}
\end{figure}

Section~\ref{subsection:fm-training-setup} identifies two elements of the training setup as important for learning the correct size distribution: placing pad atoms at the centre-of-mass (CoM), and applying a permutation alignment between the padded prior and data molecules. Here we ablate this decision by training models on GEOM Drugs only, without the pocket encoder, under an otherwise identical training setup. The first keeps both choices, the second removes the permutation alignment, and the third places pad atoms on resampled real atom positions instead of the centre-of-mass. Sampling is identical in all three cases, since the pad coordinate mode only affects the data molecule during training. We generate 1000 molecules from each model with all conditioning dropped, and take molecular size from the model's own pad token predictions, so that it is defined even where a molecule cannot be sanitised. Reported intervals are 95\% percentile bootstrap intervals over 2000 resamples of the generated molecules.

We find that the combination of CoM-padding and permutation alignment is crucial for learning the correct size distribution, although even the combined model does not perfectly match the training set sizes (Fig.~\ref{fig:size-ablation}). Against a training mean of $24.8$ heavy atoms, the combined setup generates a mean size of $22.1\,[21.8, 22.4]$, removing the permutation alignment gives $16.3\,[15.9, 16.6]$, and removing centre-of-mass padding gives $18.7\,[18.4, 18.9]$. Connected validity stays above $0.97$ and uniqueness above $0.99$ in every run, so the ablations do not degrade generation in general; they specifically move the size distribution.

The combined training setup gives pad atoms a single consistent target at the origin, which we hypothesise provides cleaner training signal. Centre-of-mass padding provides one place for them to go, and the permutation alignment assigns the prior atoms close to the origin to the pad positions, so they travel only a short distance along the trajectory. Removing either leaves a noisier signal for separating pad atoms from real atoms, and the model resolves this by padding more, which produces smaller molecules. Some undershoot remains with both choices in place, at roughly $2.6$ heavy atoms below the training mean. We attribute this to the rarity of fully unconditional samples during training, which the model sees in around 10\% of steps.

\subsection{Learning Adaptive Symmetries}
\label{appendix:adaptive-symmetry}

\begin{figure}[t]
    \centering
    \includegraphics[width=\textwidth]{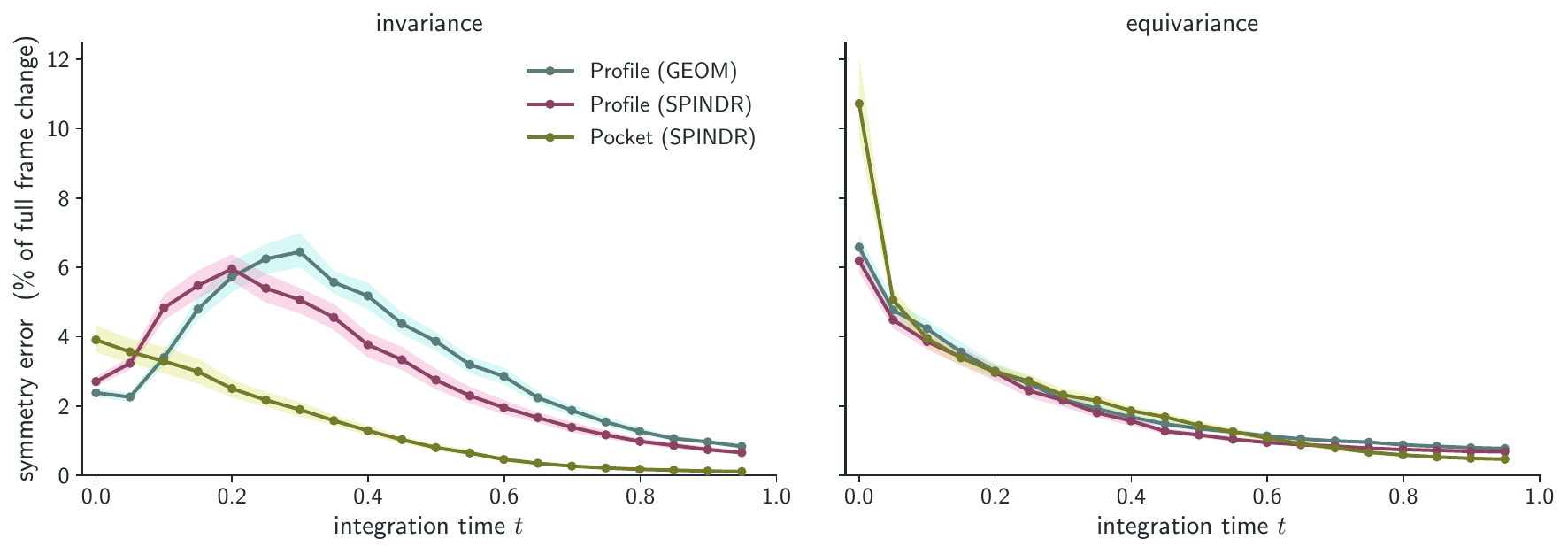}
    \caption{Symmetry error against integration time, for the profile encoder on GEOM Drugs and SPINDR ligands and for the pocket encoder on SPINDR pockets. \textbf{Left}: invariance, where the condition is rotated and the prediction should not move. \textbf{Right}: equivariance, where the state and condition are rotated together and the prediction should rotate with them. Errors are given as a percentage of the distance the prediction would have moved had it broken the symmetry instead, which ranges from $2.6$ to $5.0$Å over the encoders and snapshots shown. Bands are 95\% bootstrap intervals over 64 systems, each averaged over 8 rotations.}
    \label{fig:symmetry}
\end{figure}

Section~\ref{subsection:arch} describes adaptive symmetry learning, where the \textit{rotated} flag tells an encoder whether its input shares a reference frame with the ligand. Nothing in training constrains the encoder embeddings themselves, only the behaviour of the model as a whole, so we measure both symmetries at the output of the generator. We run two experiments, one for invariance and one for equivariance, testing the generator's predicted coordinates $\mathbf{\hat{X}}_{\theta} (.,.,.,.)$. With the flag set, rotating the condition alone should leave the prediction unchanged, such that $\mathbf{\hat{X}}_{\theta} (x_t, t, Rm, \eta) = \mathbf{\hat{X}}_{\theta}(x_t, t, m, \eta)$. With the flag unset, rotating the state and the condition together should rotate the prediction, such that $\mathbf{\hat{X}}_{\theta} (Rx_t, t, Rm, \eta) = R \mathbf{\hat{X}}_{\theta} (x_t, t, m, \eta)$. 

States are taken from the model's own generation trajectory, using constant integration steps so that snapshots are evenly spaced in $t$, with each experiment probing a trajectory generated under the flag it tests. Writing $\mathbf{\hat{X}}$ for the prediction under the unrotated condition and $\mathbf{\hat{X}_k}$ for the prediction under rotation $R_k$, we measure both RMSD$(\mathbf{\hat{X}}, \mathbf{\hat{X}_k})$ and RMSD$(R_k \mathbf{\hat{X}}, \mathbf{\hat{X}_k})$ over all atom slots. Invariance requires the first to vanish and equivariance the second, so in each experiment one distance is the symmetry error while the other gives the scale of a full frame change, and we report the error as a percentage of the latter. We use 64 systems per encoder with 8 rotations each, averaging both distances over rotations and then over systems before taking their ratio at each $t$. We normalise at each $t$ since the scale of a full frame change grows as the endpoint prediction expands from a collapsed guess at $t=0$ into a full molecule, ranging from 2.6 to 5.0 Å over the encoders and snapshots shown. Bands are 95\% percentile bootstrap intervals over systems.

Both symmetries hold and, crucially, converge towards zero as the generation proceeds (Fig.~\ref{fig:symmetry}). Invariance error stays below $6.5\%$ of a full frame change at every point and equivariance below $11\%$, the largest single value in either experiment being the pocket encoder at $t=0$. By $t=0.95$ every run is below $1\%$, with the pocket encoder the most invariant at $0.11\%$. Atom and bond distributions follow the same pattern, reaching total variation distances of at most $0.060$ and $0.006$ and falling below $0.009$ and $0.001$ by $t=0.95$. Invariance error for the profile encoder peaks part way along the trajectory, at $t=0.20$ on SPINDR ligands and $t=0.30$ on GEOM Drugs rather than at $t=0$, while the pocket encoder decreases throughout. It is not immediately clear what causes this pattern and why it is different for different encoders, but we do not attempt to investigate this further in this work. We do, however, speculate that, for generative models that rely on iterative denoising, encoding exact symmetries into architectures may be unimportant since the error in learned symmetries reduces towards zero as the molecule is resolved.

\subsection{Multi-mode Conditioning}
\label{appendix:multi-shape-cond-results}

\begin{table}[t]
    \small
    \centering
    \setlength{\tabcolsep}{4pt}
    \caption{Multi-mode shape conditioning results for different conditioning setups and values of $\sigma_{shape}$, all generated with $\gamma = 4.0$. \textit{Valid}: proportion of molecules which are RDKit sanitisable with no disconnected fragments; \textit{Unique}: proportion of unique generated molecules; $T_{\mathrm{cpt}}$ and $T_{\mathrm{ext}}$: the mean best-conformer shape Tanimoto to the compact and extended targets, respectively; $\Delta_{\mathrm{cpt}}$, $\Delta_{\mathrm{ext}}$: the mean gain over a size-matched baseline for compact and extended, respectively; and \textit{Ref. sim.}: mean ECFP tanimoto similarity to the two reference molecules. All entries are means.}
    \begin{tabular}{l c c c c c c c c}
        \toprule
        Mode & $\sigma$ & Valid & Unique & $T_{\mathrm{cpt}}$ & $T_{\mathrm{ext}}$ & $\Delta_{\mathrm{cpt}}$ & $\Delta_{\mathrm{ext}}$ & Ref. sim. \\ 
        \midrule
        \multicolumn{9}{l}{\textit{Mode targeting}} \\
        \multirow{2}{*}{Cpt (+, equiv.), Ext (+, inv.)}
         & 0.2 & 0.970 & 1.000 & 0.867 & 0.765 & 0.188 & 0.071 & 0.260 \\
         & 0.5 & 0.982 & 1.000 & 0.768 & 0.750 & 0.088 & 0.056 & 0.137 \\
        \addlinespace
        \multirow{2}{*}{Cpt (+, inv.), Ext (+, equiv.)}
         & 0.2 & 0.982 & 0.992 & 0.754 & 0.882 & 0.074 & 0.188 & 0.254 \\
         & 0.5 & 0.994 & 1.000 & 0.741 & 0.783 & 0.061 & 0.089 & 0.138 \\
        \midrule
        \multicolumn{9}{l}{\textit{Mode avoiding}} \\
        \multirow{2}{*}{Cpt (+, equiv.), Ext (-, inv.)}
         & 0.2 & 0.981 & 1.000 & 0.890 & 0.585 & 0.223 & -0.056 & 0.310 \\
         & 0.5 & 0.991 & 1.000 & 0.769 & 0.582 & 0.102 & -0.059 & 0.138 \\
        \addlinespace
        \multirow{2}{*}{Cpt (-, inv.), Ext (+, equiv.)}
         & 0.2 & 0.974 & 0.895 & 0.587 & 0.905 & -0.081 & 0.264 & 0.284 \\
         & 0.5 & 0.985 & 1.000 & 0.581 & 0.786 & -0.086 & 0.144 & 0.137 \\
        \bottomrule
    \end{tabular}
    \label{tab:full-multicond-results}
\end{table}

Table~\ref{tab:full-multicond-results} provides full results for the mode targeting and mode avoiding benchmarks of Section~\ref{subsection:multi-cond-expts}, with one row per conditioning mode and shape noise level. Alongside the shape matching gains $\Delta$ plotted in Fig.~\ref{fig:multi-cond-delta}, we report the raw shape Tanimoto achieved against each target, as well as validity, uniqueness, and similarity to the source molecules. All entries are means over the pairs in the corresponding benchmark, and $\Delta$ is measured against the size-matched baseline described in Section~\ref{subsection:multi-cond-expts}. Full definitions of each metric are provided in Appendix~\ref{appendix:eval-metrics}.

\subsection{Ensemble Property Optimisation}
\label{appendix:prop-opt-results}

\begin{table}[t]
    \small
    \centering
    \caption{Property optimisation results, all generated with $\gamma=2.0$ and $\alpha = (1)$, conditioning on the pocket and reference pharmacophores as a single mode. \textit{Valid}: proportion of molecules which are RDKit sanitisable with no disconnected fragments; \textit{Ref. sim.}: ECFP tanimoto similarity to the reference molecule; \textit{Int. rec.}: fraction of reference interactions recovered. \textit{Vina} and \textit{Vina min.} are measured in kcal/mol. All results correspond to the mean value over all benchmark systems.}
    \begin{tabular}{lcccccccc}
        \toprule
        Conditioning & Valid & Ref. sim. & Int. rec. & Vina & Vina Min. & PSA (\AA$^2$) & RMSD (\AA) \\
        \midrule
        Pocket + pharma & 0.942 & 0.191 & 0.954 & $-6.18$ & $-6.79$ & 167 & 2.07 \\
        \midrule
        + PSA = 80 & 0.939 & 0.164 & 0.945 & $-6.17$ & $-6.81$ & 111 & 2.18 \\
        + PSA = 100 & 0.935 & 0.172 & 0.944 & $-6.26$ & $-6.87$ & 129 & 2.16 \\
        + PSA = 120 & 0.922 & 0.171 & 0.954 & $-6.30$ & $-6.90$ & 144 & 2.15 \\
        + PSA = 140 & 0.922 & 0.172 & 0.958 & $-6.31$ & $-6.89$ & 163 & 2.11  \\
        \midrule
        + RMSD = 1.0 & 0.952 & 0.179 & 0.954 & $-6.90$ & $-7.39$ & 156 & 1.69 \\
        + RMSD = 1.5 & 0.953 & 0.190 & 0.961 & $-6.69$ & $-7.19$ & 159 & 1.79 \\
        + RMSD = 2.0 & 0.946 & 0.186 & 0.958 & $-6.48$ & $-7.00$ & 157 & 1.99 \\
        + RMSD = 2.5 & 0.935 & 0.174 & 0.960 & $-6.18$ & $-6.83$ & 154 & 2.17 \\
        \midrule
        Reference & 1.00 & 1.00 & 1.00 & $-7.74$ & $-7.93$ & 169 & 1.92 \\
        \bottomrule
    \end{tabular}
    \label{tab:full-propopt-results}
\end{table}

Table~\ref{tab:full-propopt-results} gives full results for the property optimisation benchmark of Section~\ref{subsection:prop-opt-expts}, with all values reported as means over the benchmark systems. Reference values are computed by passing the system's reference ligand through the same evaluation as the generated molecules; its ensemble properties come from the same MMFF sampling procedure and its Vina scores from the same in-place scoring and local minimisation. Its validity, interaction recovery and reference similarity are $1.00$ by construction, since the reference is compared against itself, and are shown only for completeness. Full definitions of each metric are provided in Appendix~\ref{appendix:eval-metrics}.

\end{document}